\documentclass[10pt,twocolumn,letterpaper]{article}

\usepackage{iccv}
\usepackage{times}
\usepackage{epsfig}
\usepackage{graphicx}
\usepackage{amsmath}
\usepackage{amssymb}
\usepackage{subfig}

\usepackage[english]{babel}
\usepackage{amsmath,amsfonts,amsthm,amssymb}
\usepackage[dvipsnames]{xcolor}
\usepackage{soul}
\usepackage{subfig}
\usepackage{graphicx}
\usepackage{rotating}  % for tables

\usepackage{pdflscape}
\usepackage{float}
\usepackage{booktabs}
\usepackage{listings}
\usepackage[toc]{appendix}
\usepackage[nottoc]{tocbibind}

\usepackage{eso-pic}
\usepackage{xspace}

\usepackage{placeins} % for float barrier

\usepackage{multirow} % table formatting

\usepackage{xcolor} % for highlighting changes

\usepackage{hanging}

\usepackage{adjustbox}      
\usepackage{algorithm}
\usepackage{algpseudocode}
\usepackage{caption}

\usepackage{comment}
\usepackage{enumitem}

\usepackage[breaklinks=true,bookmarks=false]{hyperref}

\iccvfinalcopy % *** Uncomment this line for the final submission

\def\iccvPaperID{****} % *** Enter the ICCV Paper ID here

\ificcvfinal\fi

\newcommand{\convnet}{convolutional neural network\xspace}
\newcommand{\convnets}{convolutional neural networks\xspace}
\newcommand{\stn}{STNet\xspace}

\begin{document}

%%%%%%%%% TITLE
\title{Selective Segmentation Networks Using Top-Down Attention}

\author{Mahdi Biparva, John Tsotsos\\
	Department of Electrical Engineering and Computer Science\\
	York University\\
	Toronto, Canada\\
	{\tt\small \{mhdbprv,tsotsos\}@cse.yorku.ca}}

\maketitle
\ificcvfinal\thispagestyle{empty}\fi

%%%%%%%%% ABSTRACT
\begin{abstract}
Convolutional neural networks model the transformation of the input sensory data at the bottom of a network hierarchy to the semantic information at the top of the visual hierarchy. Feedforward processing is sufficient for some object recognition tasks. Top-Down selection is potentially required in addition to the Bottom-Up feedforward pass. It can, in part, address the shortcoming of the loss of location information imposed by the hierarchical feature pyramids. We propose a unified 2-pass framework for object segmentation that augments Bottom-Up \convnets with a Top-Down selection network. 
We utilize the top-down selection gating activities to modulate the bottom-up hidden activities for segmentation predictions. We develop an end-to-end multi-task framework with loss terms satisfying task requirements at the two ends of the network. We evaluate the proposed network on benchmark datasets for semantic segmentation, and show that networks with the Top-Down selection capability outperform the baseline model. Additionally, we shed light on the superior aspects of the new segmentation paradigm and qualitatively and quantitatively support the efficiency of the novel framework over the baseline model that relies purely on parametric skip connections.
\end{abstract}

%%%%%%%%% BODY TEXT
\section{Introduction}
\label{sec:int}

In both human and machine vision systems, two directions of information
flow have been commonly considered, a data-driven or feedforward direction (Bottom-Up), and a reverse direction (Top-Down) that has a predictive,
controlling or modulatory role. In the Bottom-Up (BU) pathway, the
sensory input data is processed and sequentially transformed into
high-level semantic information such that some task criterion is satisfied
at the inference phase, while during the training phase, the error gradient
signals are calculated according to a loss function and gradients
are propagated down to the early layers for the updating of the network's
weight parameters. Convolutional neural networks model the BU pathway
for visual tasks such as object classification, detection, and segmentation.

The TD pathway, on the other hand, inherently characterizes modulatory
and controlling roles and leverages selection mechanisms. TD selection
approaches have been used for tasks such as object localization, object
segmentation, and network visualization. Selective Tuning \cite{tsotsos1995SelTun,tsotsos2011computational}
is a computational model of visual attention, and is an early attempt
to establish the applicability of a TD selection pass along with a
BU feedforward pass for basic visual tasks in dynamical networks.
Selective Tuning of \convnets (STNet) \cite{biparva2017stnet} has proposed to formulate a neural network framework with a selective TD mechanism and is evaluated for the object localization task. STNet validates the effective role of the TD selection pass to localize relevant regions covering the object features.
In this work, we investigate the role of TD selective attention in \convnets and the feature modulation of BU hidden activities for the task of object segmentation. We attempt to complement the STNet-for-localization formulation in this work with feature modulation of BU hidden activities. This work strives to examine whether feature modulation derived from a TD selective pass is successful for the demanding object segmentation task. 

% Why modulation
The purpose behind TD feature modulation is to select and modify the data interpretations represented by the hidden BU activities.
%of the BU hidden activities is that the TD gating activities are merely selective in nature while the BU hidden activities are representative. 
Therefore, for a subsequent processing stage such as object segmentation, the TD selection patterns modulate the BU feature encoding activities. As a result, the subsequent segmentation process will benefit from both of the densely-encoded bottom-up flow of information and the sparsely-selective top-down patterns of modulation.

Dominant approaches for object segmentation are generally densely parametric in a fully-convolutional manner. Multiple up-sampling convolutional layers are leveraged to predict up-scaled category maps at the final segmentation layer. Approaches such as skip connections, multi-level feature augmentation, and discriminative attention are proposed to produce fine-grained segmentation. All such approaches enforce up-sampling of predicted score maps of a pre-trained network through a number of parametric layers in a purely feedforward manner. While they have reached a promising performance level, we show the up-sampling part does not need to be densely parametric. Rather, a successful hierarchical TD selection through the BU network can be helpful to modulate rich feature maps for segmentation map predictions.

We propose the Selective Segmentation Network (SSN) to systematically
study the use of the TD selection pass for the object segmentation
task. SSN consists of the BU pass that utilizes a typical \convnet
with multiple layers of feature extraction. To deal with the multi-instance and multi-scale issues in the experimental evaluation, we define a controlling module called Loose Spatial Detection (LSD) that examines high-level semantic information
and loosely predicts the locations and scales at which TD attention needs to be activated. Once important locations and scales
are determined, attention signals are set and the TD selection pass is
activated. The TD selection mechanism has three stages of processing at each
layer that relies on two stages of local competitions and one stage
of normalization for gating activity propagation. Gating activities
at each layer are computed and the selection is passed to the lower
layer until some early layer in the visual hierarchy is reached.

The TD pass systematically computes selection patterns over relevant
hidden features of the BU network. We develop our investigation around
the hypothesis that the TD selection patterns are reliable
as a source of hidden feature modulation for segmentation. We propose
to have the information flow of hidden feature activities modulated
by the TD gating activities into the segmentation pipeline for the
final output predictions. We study the effect of three types of
modulation at different stages of computation on the prediction performance
of SSN. 
The segmentation pipeline forms a cascade of parametric blocks each consisting of a number of processing units. Each block performs operations such as input feature modulation, channel reduction, and parametric feature fusion.
At each layer the modulated input information from the BU and TD passes
is integrated into the segmentation pipeline and then using parametric transformation is passed to the layer below. After a few layers, the final segmentation maps are predicted at the bottom of the visual hierarchy.

SSN benefits from a multi-task formulation. We define two loss functions: one at the top of the visual hierarchy where the LSD module outputs the label predictions for attention signal initialization and one at the bottom of the visual hierarchy where the segmentation output maps are generated. 
The former loss measures the capability of the BU network and LSD module to jointly predict the starting points for TD pass initialization.
The latter loss measures the performance of SSN to predict segmentation
maps. 
%The final loss function is formed by a weighted addition of the two loss functions.

Learning in SSN has two phases: in the first phase, the pre-trained parameters of the BU network and the uniformly initialized parameters of the LSD modules are jointly fine-tuned using the first loss function before being loaded into the complete SSN framework. In the second phase, SSN is loaded with the parameters obtained in the first phase, and the entire segmentation network is trained with the two loss functions using the Stochastic Gradient Descent (SGD) algorithm.
%Then TD selection modulates information, and finally the segmentation block predicts output maps. We optimize the SSN loss function using SGD algorithm.

SSN is qualitatively and quantitatively evaluated on the visual task of semantic segmentation. Semantic segmentation \cite{chen2018deeplab,everingham2015pascal,LinMaireBelongieEtAl2014,cordts2016cityscapes} is the task of predicting pixel-level category labels given a pre-defined list of object classes. Unlike the object localization task that outputs a number of bounding boxes enclosing category instances, semantic segmentation returns a segmentation map containing a category label at each pixel.

We illustrate that SSN improves the baseline model for the evaluation performance on three benchmark datasets: Pascal VOC, CamVid, and Horse-Cow datasets. 
We conduct ablation studies to shed light on aspects of feature modulation using TD selection such as feature entanglement and noise interference. 
Experimental results reveal that the modulatory nature of the TD pass helps to untangle the underlying feature representations to some degree and improves the results of the segmentation metrics under input perturbation scenarios such as additive noise and box occlusion.

\section{Related Work}
\label{sec:rel}
There are two main classes of previous research that are related to our own, and a brief overview will be provided under the headings of Semantic Segmentation and Top-Down Approaches.

%Convolutional neural networks have been successful in recent years on various visual tasks. They are trained and evaluated on benchmark datasets. 
\textbf{Semantic Segmentation:} The Fully-Convolutional
Network (FCN) \cite{long2015fully} has been a pioneer to introduce
convolutional encoder-decoder networks that benefit from parametric
skip connections and fractionally-strided convolutions to gradually
up-sample in a parametric fashion the output of the label prediction
layer at the top of a regular \convnet. The architecture implements an information bottleneck using the encoding network which is basically an extension of a multi-layer classifier and the decoding network that up-samples the semantic label predictions of the encoding network into the output segmentation map. The FCN approach has been extended
with novel approaches for performance improvements \cite{noh2015learning,chen2015semantic,wang2016objectness,chen2016attention,chen2018deeplab,islam2017gated}.
These approaches are mainly involved with modifications and extensions such as novel network architectures (\eg residual networks \cite{he2016deep}), addition of extra parametric layers and dense connectivity, multi-scale
augmentation, and multi-level supervision to improve the segmentation prediction accuracy specially for small and fine-detailed objects. These approaches have shown success in terms of the evaluation performance metrics. We attempt to study the complementary role of Top-Down selection to such encoder-decoder frameworks. Furthermore, unlike the FCN model, the skip connections are no longer densely merged into the decoding segmentation pipeline. The modulation of hidden activities using the TD activities is introduced in this work in the hope of achieving higher performance results and robustness to out-of-distribution input perturbations.

\section{Selective Segmentation Network}
\label{sec:mod}
\begin{figure}
	\includegraphics[width=0.9\columnwidth]{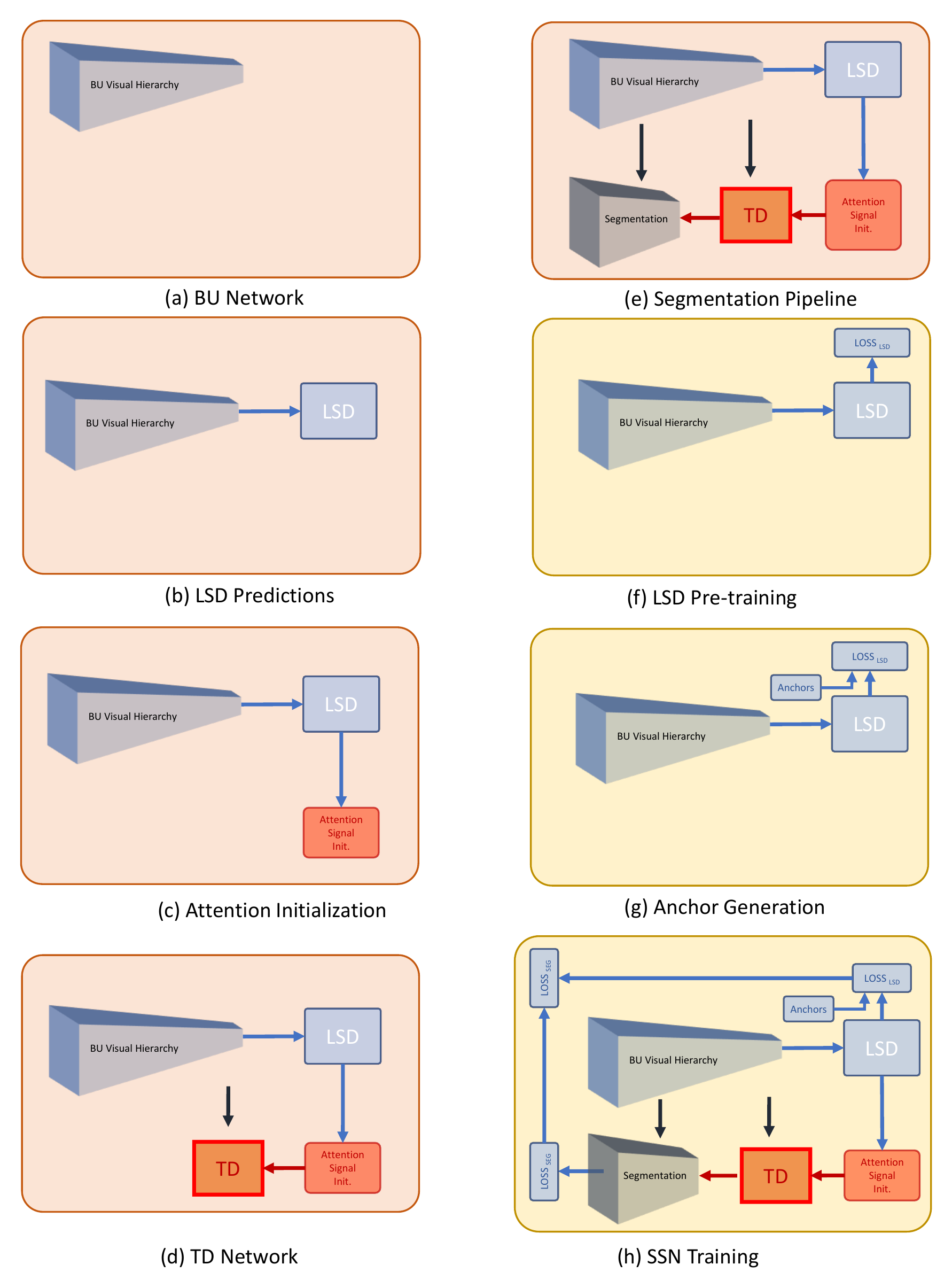}
	\caption{\label{fig:ssn-overall-modular}Illustration of the modular information flow of the Selective Segmentation Network (SSN) at each processing stage of the inference and learning phases. The stages in orange belong to the inference phase at which given some unknown test image, the predicted segmentation outputs are returned. The stages in yellow represent the learning phase at which SSN parameters are learned. The text provides details for each of the figure panels.}
\end{figure}

The Selective Segmentation Network (SSN) consists of three major processing units: the BU network, the TD network, and the segmentation network:
1) the BU representation is the core visual hierarchy that consists
of multiple parametric layers, 2) the TD selection mechanism produces gating activities using attentional traces throughout the visual hierarchy, and 3) the attentive segmentation network modulates hidden activities with gating activities at a number of different levels and merges them into a unified representation with gradual up-sampling for the final segmentation prediction. We are going to provide a procedural model overview in Sec. \ref{sec:seg:model:overview} explaining different stages of processing in SSN at the inference and learning phases briefly. In the subsequent sections, each stage will be explained with mathematical formulation and implementation details.

\begin{figure*}[t]
	\includegraphics[width=1\textwidth]{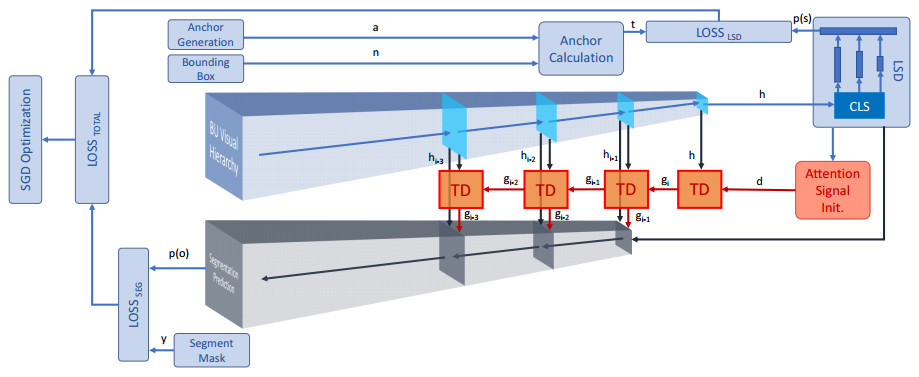}
	\caption{Illustration of SSN consisting of multiple parts such as the feedforward
		BU representation, the classification LSD module, the TD selection
		network, and the up-sampling segmentation pipeline. Arrows show the
		information flow from one part to another part at the learning phase. The input and output at each stage are labeled using the variables which are defined in the subsequent sections.
		\label{fig:ssn-overall-architecture}}
\end{figure*}

\subsection{Method Overview}
\label{sec:seg:model:overview}

Fig. \ref{fig:ssn-overall-modular} demonstrates the computational stages of SSN at the inference and learning phases. We first begin with the stages involved in the inference phase and then move on to the learning phase. In the inference phase the information flows into SSN sequentially as follows. In Fig. \ref{fig:ssn-overall-modular} (a), the BU feedforward feature representation is defined by a \convnet. The input image is transformed by multiple feature extraction layers into semantic information for some particular task prediction. Details are given in Sec. \ref{sec:seg:model:Bottom-Up-Feature-Encoding}. In Fig. \ref{fig:ssn-overall-modular} (b), the LSD module is defined to deal with the multi-instance and multi-scale issues in semantic segmentation. The objective of LSD is to predict the top output units at which TD selection mechanisms need to be activated. Details are given in Sec. \ref{sec:seg:model:Loose-Spatial-Detection}. In Fig. \ref{fig:ssn-overall-modular} (c), based on the prediction scores returned by LSD, the attention signal initialization unit determines the positions and scales to which attention must be deployed. We propose three different initialization strategies for which the details are given in Sec. \ref{sec:seg:model:Attention-Initialization}. In Fig. \ref{fig:ssn-overall-modular} (d), TD selection begins from the initialization signal and traverses downward in a layer-by-layer manner. TD selection at each layer produces gating activities representing feature importance across spatial positions and channels. They influence the flow of hidden activities into the segmentation network. Details are given in Sec. \ref{sec:seg:model:Top-Down-Selection}. In Fig. \ref{fig:ssn-overall-modular} (e), the last stage of the inference phase is the segmentation prediction. At this point, both of the BU hidden and TD gating activities are produced for the input image and are the input to the segmentation pipeline. It has multiple levels of feature modulation, channel reduction, feature fusion, and spatial up-sampling. Further details are given in Sec. \ref{sec:seg:model:Segmentation-Prediction}.

In the learning phase, we deal with the optimization of the SSN parameters for semantic segmentation given the new input data domain and task requirements. In the first learning stage as depicted in Fig. \ref{fig:ssn-overall-modular} (f), LSD pre-training is defined to adapt the feature representation of the BU network according to the LSD layers prior to the segmentation stage. In Fig. \ref{fig:ssn-overall-modular} (g), in order to optimize the BU and LSD parameters, proper target variables need to be produced from the provided ground truth bounding boxes. The LSD pre-training and anchor generation are explained in detail in Sec. \ref{sec:seg:model:LSD-Pre-training}. In Fig. \ref{fig:ssn-overall-modular} (h), the full SSN model is trained in the multi-loss setting using the LSD and segmentation loss functions. The former sits at the top of the hierarchy while the latter is at the bottom. The ground truth segmentation mask is the target variable used for the segmentation loss function. Details for the multi-loss setting are given Sec. \ref{sec:seg:model:Multi-loss-Training}.

Fig. \ref{fig:ssn-overall-architecture} illustrates different parts of SSN in the learning phase all together, the information flow from one processing unit to another one, and the outputs at the top and the bottom of the hierarchy in more details. The input and output at each stage are labeled using the notations developed in the subsequent sections. It also depicts the joint loss function for the training of the entire network in an end-to-end manner using the SGD optimization algorithm. In the following, we explain the sequence of computational stages for the inference and learning phases in more detail.

\subsection{Bottom-Up Feature Encoding}
\label{sec:seg:model:Bottom-Up-Feature-Encoding}
%\textbf{Definition:}
Information processing in SSN begins with the BU network that encodes the low-level input sensory data into high-level output semantic information at the top of the network. The BU network consists of multiple layers of feature extraction such as convolutional layers, non-linear transfer functions, and pooling layers. The spatial resolutions of the output maps throughout the network are gradually decreased while the feature channel size is increased. BU layers are defined according to a pre-defined network architecture such as AlexNet \cite{krizhevsky2012imagenet} or VGG-16 \cite{simonyan2013deep}. Part (a) of Fig. \ref{fig:ssn-overall-modular} illustrates the BU feature encoding as the first stage in the inference phase.

%\textbf{Formulation:}
The training set $D=\{(x_{i},y_{i})\}_{i=1}^{N}$ contains $N$ samples
such that each sample consists of an input image $x\in\mathbb{R}^{3\times H\times W}$
and the ground truth $y$. The ground truth $y=(y^B, y^S)$ contains the bounding box annotations $y^B$ of the category instances in the input image and the segmentation target mask $y^S\in\mathbb{R}^{1\times H\times W}$, where $H$ is the input image height, $W$ is the input image width. A pixel element on the segmentation mask at the vertical and horizontal position $(h, w)$ has a category label $y^B_{hw}\in\{0,1,\dots,K-1\}$ for $K$ different category labels including
the background label $0$. The BU pass consists of a multi-layer feedforward
convolutional network

\begin{equation}
h=f(x;W_{BU}),
\end{equation}
where $f$ is a cascade of neural network layers, such as convolutional
and pooling layers, and is parameterized with the set of connection
weights $W_{BU}$ depending on the underlying network architecture,
$x$ is the input image to the network, and $h\in\mathbb{R}^{C_{f}\times H_{f}\times W_{f}}$
is the hidden activity map at the top of the network with feature
channel size $C_{f}$ and spatial size $H_{f}\times W_{f}$. In a nutshell, the BU pass gradually transforms the raw input data using a number of parametric layers into a high-level semantic information for label predictions. 

\begin{figure}
	\includegraphics[width=0.9\columnwidth]{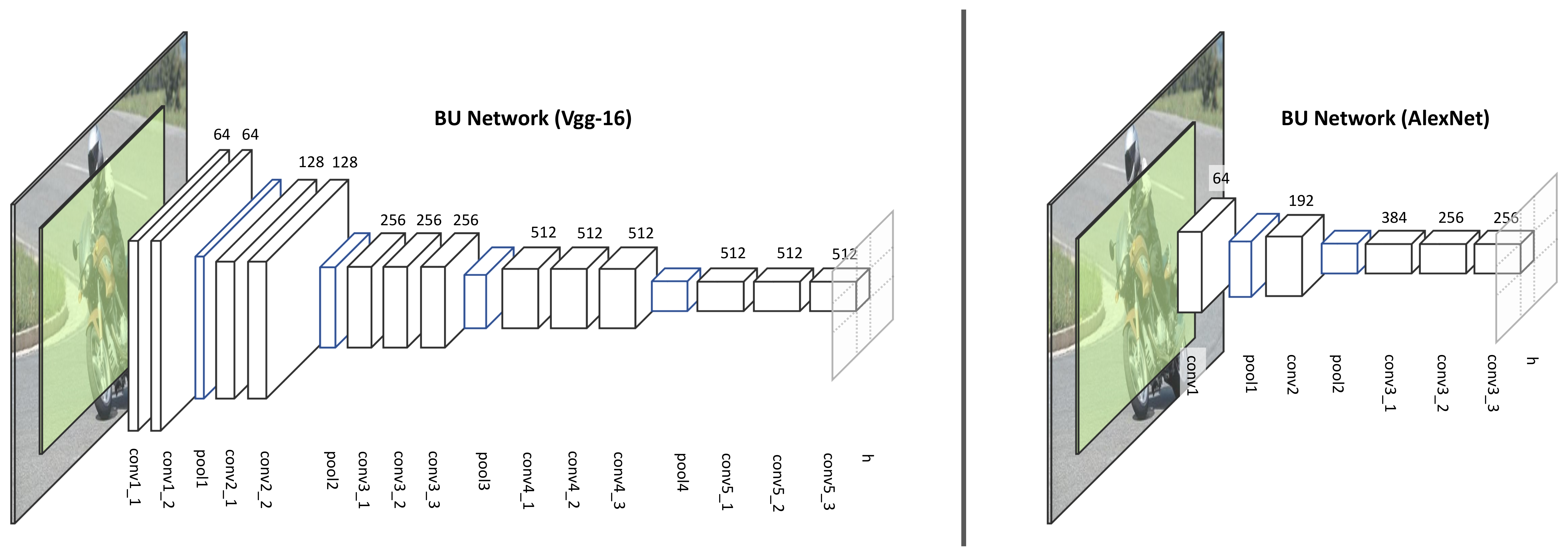}
	\caption{\label{fig:ssn-bu-archs}The BU network defined using the AlexNet and VGG-16 \convnet architectures on the right and left respectively. The green box over the input image is the total receptive field size of a unit on the top feature map $h$. Blue boxes are pooling layers and the black boxes are convolutional layers with ReLU activation functions. Since the total receptive field size is smaller than the input image size, the top feature maps have size of greater than 1.}
\end{figure}

\subsection{Loose Spatial Detection}
\label{sec:seg:model:Loose-Spatial-Detection}

% 00 - Motivation
%\textbf{Motivation and Objective:}
The BU network is initially loaded with a network trained for object classification on Imagenet benchmark dataset \cite{Russakovsky2015}.  The definition of object classification is to recognize one single category instance in the input image. Thus, there is always one instance of one category in the input image. Consequently, classification models are required to return one single label output for an input image. The output unit apparently has a total receptive field as large as the input image size so then the entire image is covered.
In object detection \cite{Russakovsky2015} and semantic segmentation \cite{everingham2015pascal}, on the other hand, this is no longer the case and the input image is defined to have larger spatial size and may contain multiple instances of different object categories at various spatial locations and scales. Namely, the input data domain in these two tasks has multi-instance and multi-scale characteristics. As a result, the multi-instance and multi-scale aspects must be addressed by detection and segmentation models. Object detection approaches such as FRCNN \cite{ren2015faster} and SSD \cite{liu2016ssd} devise sliding-window approaches on the feature embedding space at the top of the visual hierarchy to produce label predictions at all possible spatial positions. The multi-scale issue in SSD \cite{liu2016ssd} is addressed by defining multiple classification output layers at different levels of the visual hierarchy such that each has a wider receptive field and consequently covers a larger portion of the input image and is capable of predicting larger objects. 

\begin{figure}[t]
	\includegraphics[width=1\linewidth]{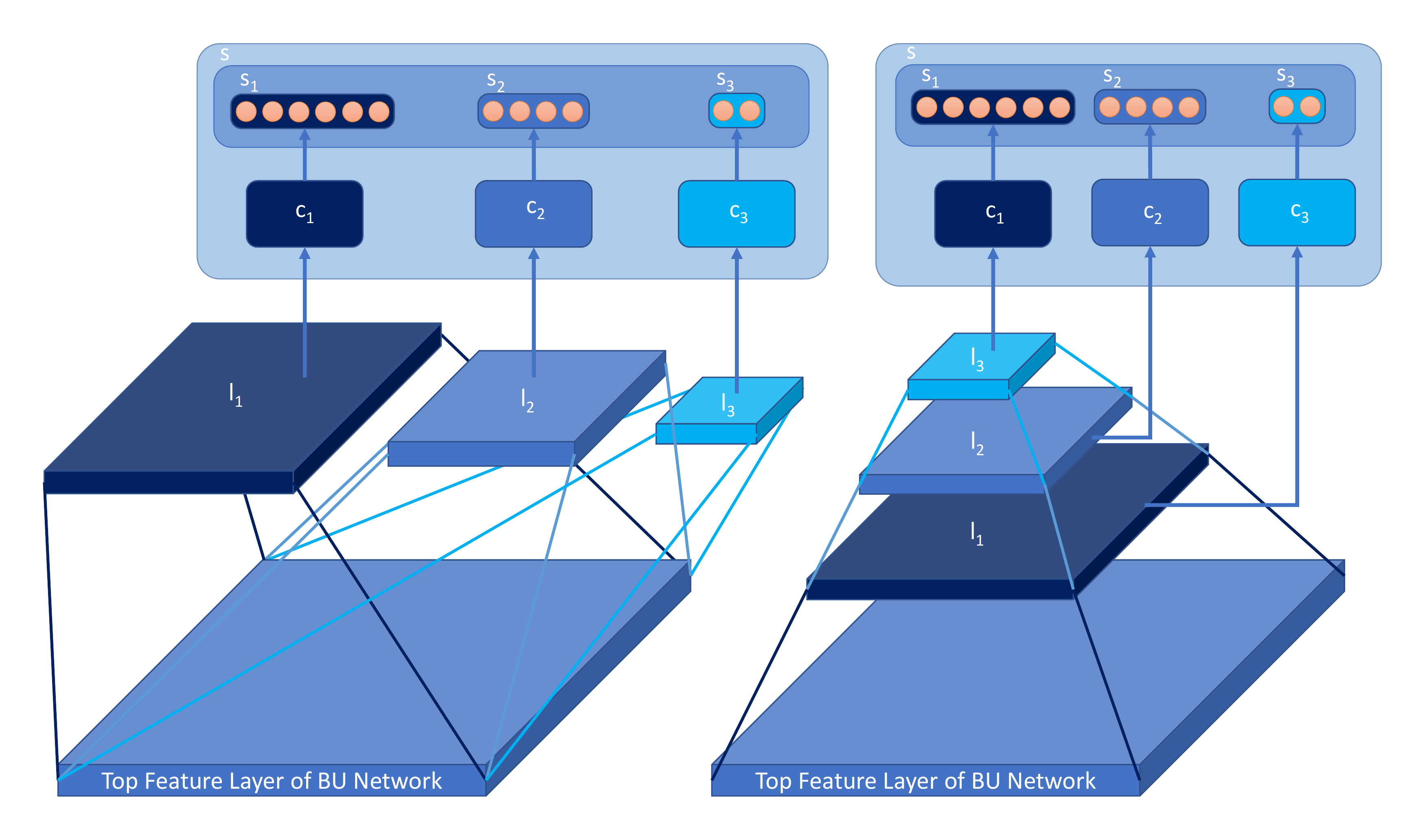}
	
	\caption{Parallel (right) and Sequential (left) architecture approaches to
		design the Loose Spatial Detection (LSD) module. Each shade of blue represents
		a group of layers with the intermediate layers $l_i$, the output prediction layers $c_i$ and the output score maps $s_i$. The top feature layer is the last layer of the BU network that outputs the feature maps $h$. The layer connectivity of the parallel and sequential choices along with the spatial size reduction from one group to another is depicted schematically.
		\label{fig:lsd-module}}
\end{figure}

In this work, we also need to address the following aspects of the input data for semantic segmentation. SSN needs to be able to trigger TD selection for the positions and scales at which there are category instances.
Following the same modeling approach as in SSD, we propose to deal with the multi-instance and multi-scale characteristics in semantic segmentation using the Loose Spatial Detection (LSD) module. It triggers the activation of the TD network at different positions and scales. LSD is a controlling unit that determines whether TD selection needs to start from an output unit at a particular position and scale. 

\begin{figure}[t]
	\centering
	\includegraphics[width=0.75\linewidth]{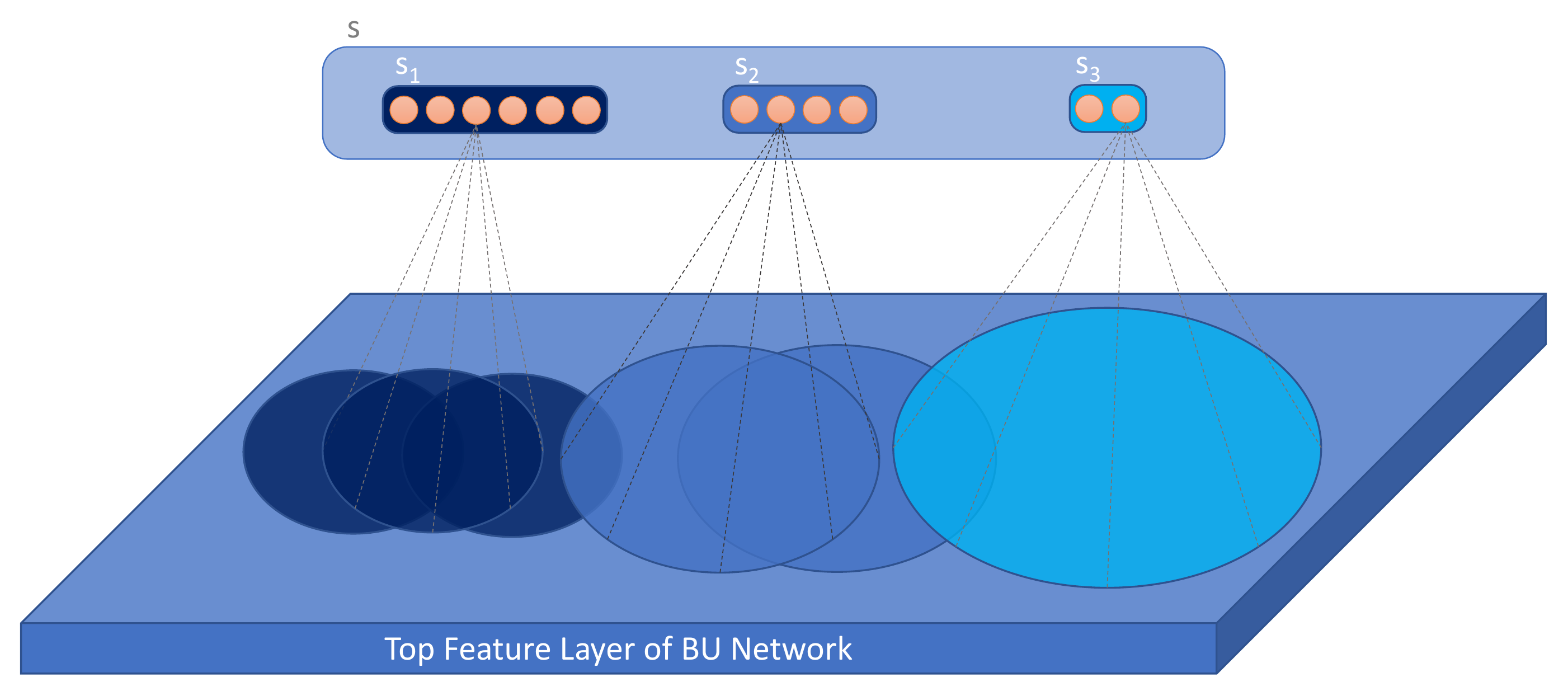}
	
	\caption{The receptive field size of three LSD groups over the input feature map $h$. The shades of blue represent the receptive field and the output score map of a particular group of LSD layer. 
		%	$H_f$, $W_f$, and $C_f$ indicate the height and width and channel size of the feature hidden maps $h$ returned by the BU network.
		\label{fig:lsd-module-rf}}
\end{figure}

% 01 - Definition: Introduction
%\textbf{Definition:}
LSD contains $C$ groups of parametric layers. In each group, there are a number of convolutional and pooling layers and the last output prediction layer has a particular total receptive field size and output spatial size. 
The total receptive field size of a layer is a spatial span in the input image space that a unit on the output maps of the layer covers. 
The output of a layer has a 2D spatial size which is calculated according to the hyperparameter settings of the layer.
%The spatial size of the output of a layer is the number of units ordered in 2-dimension and is calculated according to the hyperparameter settings of the layer. 
Settings such as kernel filter size, marginal padding of the input activities to the layer, and sub-sampling rate (stride) specifies the spatial size of the output map. 
For instance, a layer with kernel filter size 5 has a wider total receptive field size in comparison to the kernel size 3. 
We define the combination and ordering of LSD layers in each group such that the total receptive field size of the units on the output prediction maps increases from the first group to the next while the output map spatial size decreases respectively. This is achieved using a combination of convolutional and pooling layers with appropriate kernel size, stride, and padding values. Fig. \ref{fig:lsd-module-rf} schematically demonstrates the outputs of an LSD with three groups such that an node in the output score map $s_1$ has a smaller receptive field size while the number of nodes is larger. As we move to the next two groups, the receptive field size increases and the number of nodes in the output maps decreases. As illustrated in Fig. \ref{fig:ssn-overall-modular} (b), the LSD module is used right after the end of the BU pass of information processing in the second stage of the inference phase.

% 02- Mathematical Formulation
%\textbf{Formulation:}
$h$, the output of the final BU layer at the top of BU network, is fed into LSD

\begin{equation}
s=c(h;W_{LSD}),
\end{equation}
where $s\in\mathbb{R}^{K\times A}$ is the output score map
with $A$ units such that at each unit, predictions for $K$ category labels are produced, and $W_{LSD}$ is the set of LSD weight parameters. 
LSD module $c(h;W_{LSD})=\{(l_{i},c_{i})\}_{i=0}^{C}$ adds $C$ extra groups
of layers on top of the BU network where $l_{i}$ is the set of intermediate parametric layers in group $i$ and $c_{i}$ is the final output prediction (classification) layer returning the output score map $s$ such that
$s=\{s_{i}\}_{i=0}^{C},s\in\mathbb{R}^{K\times A},A=\sum_{i=0}^{C}|s_{i}|$. $|s_{i}|$ is the total number of output units returned by the classification layer $c_i$. We refer to the last output layers in LSD as output prediction, discrimination, or classification layers interchangeably.

We propose to experiment with two possible approaches to define the connectivity of layers in the LSD module as illustrated in Fig. \ref{fig:lsd-module}: the sequential and parallel architecture designs depicted in the right and left parts respectively.

As the name of the two design choices imply, the LSD output predictions are computed using a combination of parallel or sequential groups of layers. 
In the former, the $C$ groups process the input feature maps $h$ in a disjoint and parallel manner while in the latter, a group with a larger receptive field sits on top of the other with a smaller receptive field. 
Additionally, in the sequential case the parametric feature representation is shared among all of the other underlying groups by passing the output of one group as the input to the next one.
In the parallel design, on the other hand, each group maintains a separate feature representation on top of the input feature maps $h$ to produce the output predictions. So the feature representation throughout one group is not shared with the layers in another group.
%For the tasks such as object detection and segmentation, there is
%always the issue of dealing with large visual context with multiple
%objects of different scales. In SSN, we define the Loose Spatial Detection
%(LSD) module to learn to predict the locations and scales at which
%TD selection is required. The ultimate goal of the LSD module is to predict label categories at multiple scales using the expansion of the receptive field sizes by a combination of convolutional and pooling layers. 

%We elaborate more on the details of each choice in the following. 
In the sequential design, the set of intermediate layers $l_{i}$ not only pass information to the set of intermediate layers $l_{i+1}$ of the next group but also feeds into the classification layer $c_{i}$ to output label score maps $s_{i}$. Each unit in $s_{i}$ returns the confidence scores for $K+1$ category labels. The category with the highest score is basically the category for which the TD selection needs to begin at this unit.
Each set of intermediate layers $l_{i}=\{u_{i},o_{i}\}$ contains a convolutional layer $u_{i}$ with kernel size $1\times1$ followed by a convolutional layer $o_{i}$ with kernel size $3\times3$.
In the parallel design, on the other hand, each set $l_{i}$ only feeds into the classification layer $c_{i}$ and contains a number of convolutional and pooling layers.

%Fig. \ref{fig:lsd-module} schematically illustrates the parallel and sequential design choices on the left and right respectively. 
The parallel and sequential LSD types are schematically demonstrated in Fig. \ref{fig:lsd-module} for three groups of layers.
The output $h$ of the top feature layer in the BU networks is fed into the groups of layers. Boxes in different shades of blue represent the set of intermediate layers $l_i$ for $i=\{0, 1, 2\}$. They output information into the final prediction layers $c_i$ for the prediction of the score maps $s$. The output maps $s_i$ are returned by the prediction layer $c_i$ such that $|s_i| < |s_{i+1}|$ the output score map size of the first group is smaller than the second and the second smaller than the third group.

\subsection{Attention Initialization}
\label{sec:seg:model:Attention-Initialization}

%\textbf{Definition:}
Once the LSD module is finished computing the output prediction tensor $s$, we need to determine the set of elements for which the TD selection mechanisms need to be activated in the third stage of the inference phase as illustrated in part (c) of Fig. \ref{fig:ssn-overall-modular}. We propose to experiment with three different initialization strategies described in the following. The attention initialization module receives the LSD output tensor $s$ and produces an initialization signal according to one of the three strategies.
%\textbf{Formulation:}
The TD selection pass is initialized by an input attention signal $d$
such that $d=\{d_{i}|d_{i}\in\mathbb{R}^{K}\}_{i=0}^{A}$. 
$d$ contains the same number of elements as $s$ does and is initially a tensor of zero elements.
The initialization strategy determines the category for which the TD selection mechanism will be activated for a particular element $d_i$. This is achieved by the one-hot encoding representation described as follows. There might be elements for which there is no TD selection activated.

\textbf{Ground Truth Strategy:} This strategy sets the elements of the attention signal $d$ to one according to the ground truth anchor labels which are used for LSD prediction training described in \ref{sec:seg:model:LSD-Pre-training}:

\begin{equation}
	d^{GT}=\{d_{ij}=1|j=t_{i}\}_{i=0}^{A},
\end{equation}
where $t_{i}$ holds a category label for which the anchor $i$ is
determined to have the highest IoU with a ground truth bounding
box of an object of the category. The goal of this strategy is to measure the performance
of the LSD module to activate the TD pass according to the ground truth target values rather than the LSD output confidence scores.

\begin{figure}[t]
	\includegraphics[width=1\linewidth]{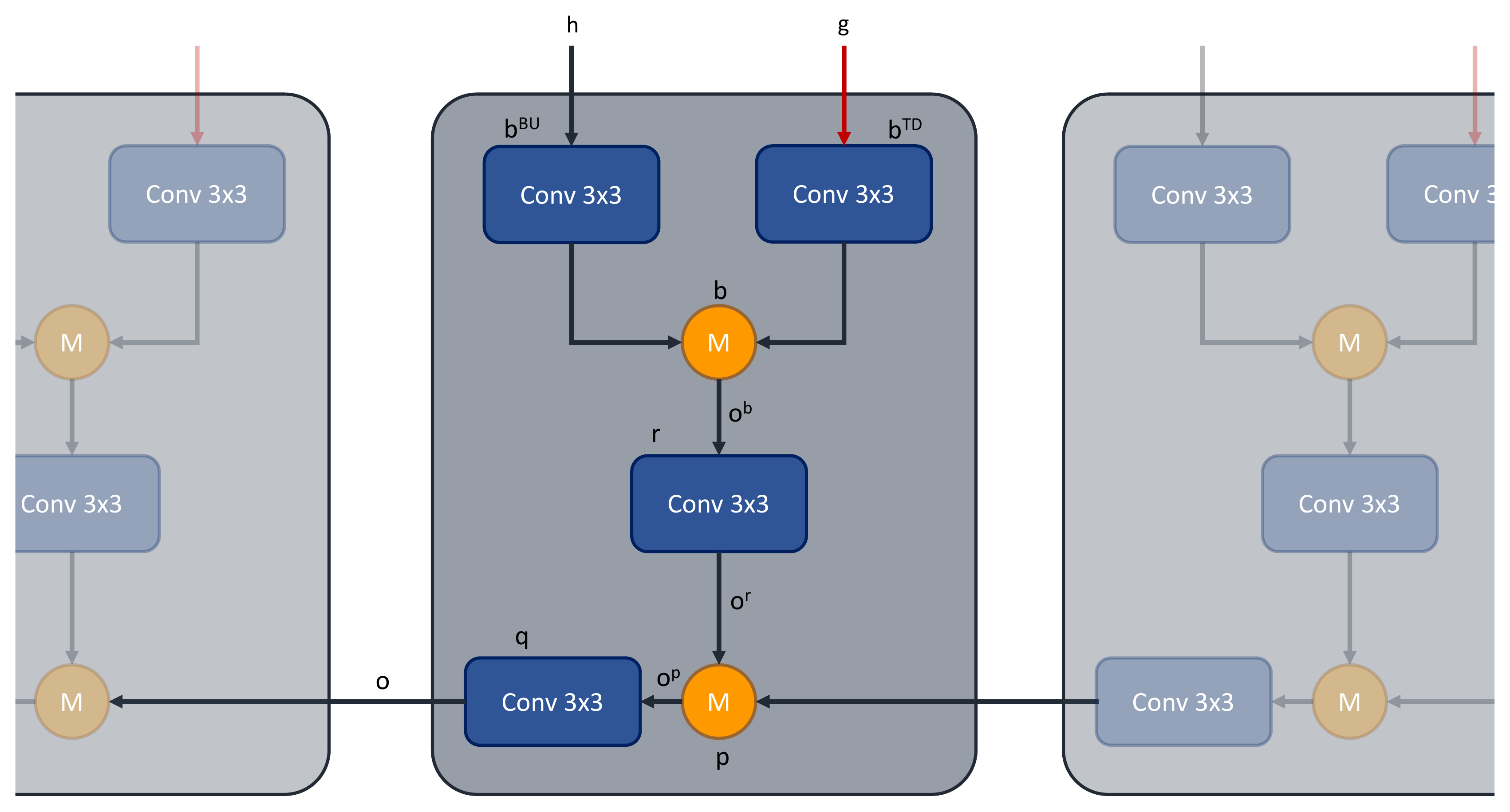}
	
	\caption{Illustration of the segmentation network with different parametric and modulation nodes.
		Each block receives the hidden (blue) and the gating (red) activity inputs. The selective gating units modulate the hidden units at the first node $M$. At each layer, after input fusion, information is integrated into the main segmentation pipeline using the second modulation node $M$. We conduct experiments on three different types of modulations: addition, multiplication, and concatenation. The layer label subscript $i$ is neglected for the sake of brevity.  \label{fig:segmentation-module}}
\end{figure}

\textbf{Top-1 Strategy:} This strategy is the most straightforward
approach to initialize the attention signal. It finds the category label of the maximum
output score value and then set the signal value for the category label to one:

\begin{equation}
	d^{top-1}=\{d_{ij}=1|j=\text{argmax}_{k}s_{ik}\}_{i=0}^{A},
\end{equation}
in which $s_{i}\in\mathbb{\mathbb{R}}^{K}$ is the LSD
score element returned by the LSD module. The reliability
of LSD is verified when the final segmentation performance of SSN
initialized using this strategy is close to the ground truth strategy.
It implies LSD has learned to determine the units for which
TD selection is essential to be activated so then the segmentation network benefits from a rich set of modulated features.

\textbf{Thresholding Strategy:} There is no purpose in imposing the TD selection process in spatial regions where there is little confidence that a target is present (this would be a false alarm). Therefore, in order to reduce redundant TD selection imposed by false alarms, we
%In order to reduce the false alarm negative impact on the segmentation performance induced by redundant TD selection mechanisms, we further define to 
threshold the maximum confidence scores using a cross-validated thresholding value $\theta^{attention}$:

\begin{equation}
	d^{\theta}=\{d_{ij}=1|j=\text{argmax}_{k}s_{ik},s_{ij}>\theta^{attention}\}_{i=0}^{A}.
\end{equation}

It reduces the redundancy in TD pass and consequently lowers the interference imposed by misleading noisy features for the segmentation pipeline. Additionally, this strategy reduces the processing time required for the completion of the TD selection pass and hence the overall SSN processing time decreases. 
%The intuition is that it removes false alarm loose detection cases,
%and consequently ignores redundant and interfering TD selection traversals.
%\textbf{Implementation Details:}
$s$ contains the probability confidence values ranging from zero to one. We cross-validate a range of thresholding value $\theta^{attention}$ and find that $\theta^{attention}=0.9$ is tight enough to improve the segmentation accuracy and maintain the TD selectivity. This value is used during the experimental evaluation.

\subsection{Top-Down Selection}
\label{sec:seg:model:Top-Down-Selection}

%\textbf{Definition:}
% taken from the method overview
The TD network computes the selection patterns through which the gating of the BU activities into the segmentation network is performed. The TD selection mechanisms are activated using the attention signal initialization module. Once the initialization signal tensor is set, the TD selection network starts processing the BU hidden activities to compute the gating activities at each layer and then the selection is passed to the layer below. This process continues until the TD pass stops at some early layer of the visual hierarchy. 
The information flow in the inference phase from the initialization module to the TD network is illustrated in part (d) of Fig. \ref{fig:ssn-overall-modular}. Further details are given in Fig. \ref{fig:ssn-overall-architecture} by depicting the TD selection mechanism at each layer and the information flow from one layer to another layer in the TD network. The gating information flows from the top layers to the intermediate and early layers in the TD pass.
It is shown in STNet model \cite{biparva2017stnet} that the TD gating
activities are sufficiently representative for object localization, and we hypothesize that they are reliable to select features for object segmentation. We experimentally support the hypothesis and show that the gating activities indeed improve the segmentation accuracy over the baseline model without a similar gating mechanism.

%\textbf{Formulation:}
We follow STNet \cite{biparva2017stnet} formulation for the TD selection
pass. The TD pass begins from the elements $d_i$, which are set to one by the initialization module. Those that are zero will not participate in the TD traversal.
%the units and at the categories for which the
%input attention signal is set to one and layer by layer traverses down until it stop at some early layer.
We define the TD network as

\begin{equation}
	g=n(d,h,W_{BU},W_{LSD}),
\end{equation}
in which $n=\{n_{i}\}_{i=J}^{V}\,|\,g_{i}=n_{i}(g_{i+1},h_{i},w_{i})$
is a set of sequential TD layers called one after each other, $d$
is the input attention signal, $h$ is the set of the BU hidden activities at all layers,
and $W_{BU}$ and $W_{LSD}$ are the set of kernel parameters of the BU network and the LSD module respectively. At every TD layer, $g_{i+1}$
and $g_{i}$ are the input and output gating maps respectively,
$h_{i}$ is the hidden activity map passed from the BU
layer $i$, and $w_{i}$ is the kernel filter weights of the BU convolutional layer. $J$ is the penultimate layer in the visual hierarchy such
that $g_{J+1}=d,h_{J}=s$ and $V$ is the level at which the TD
pass ends.

The selection mechanism $n_{i}$ is implemented by three computational stages. All
three stages are performed in the local scope of the receptive field
of a node. The computation in the stages is based on the element-wise multiplication
of the hidden activities falling inside the receptive field and the
kernel filter weights. We call this set Post-Synaptic (PS) activities
hereafter. 
The first stage takes care of noise interference reduction
by running a competition among PS activities, and determining the
set of winners. It has an adaptive thresholding mechanism to implement a local competition between PS activities. 
The second stage performs grouping and selection of
the winners according to spatial and statistical criteria for the convolutional and fully-connected layers. 
Lastly, the third stage normalizes PS activities of the selected group such
that they sum to one and propagates the gating activity proportional
to the normalized PS values to the localized gating units in the layer
below. The gating maps at each layer represent the selection patterns
that will be used for feature modulation in the segmentation pipeline.

\subsection{Segmentation Prediction}
\label{sec:seg:model:Segmentation-Prediction}

%\textbf{Definition:}
% taken from the method overview
The BU representation and TD selection integrate into a unified pipeline
in the segmentation network. 
The segmentation network is defined to learn the spatial and feature correlations of the modulated feature activities by parametric up-sampling of the feature planes for the final segmentation predictions.
Hidden activities in the BU layers represent
input sensory data according to an optimization policy such that
an objective loss function is minimized. However, the spatial resolution is reduced along the visual hierarchy due to the gradual increase
in the receptive field sizes and sub-sampling rates. Rather than adding parametric sub-sampling layers in a brute-force manner to generate
segmentation predictions similar to FCN-based approaches, TD selection
fills the gap between the spatial acuity and semantic richness by computing
selective gating activities at each layer of the hierarchy. These
activities are used to modulate hidden activities in the
spatial and feature dimensions at multiple levels. 
%The levels of each network architecture that participate in the attentive segmentation will be experimentally evaluated for the best performance result. 
%The building components of the attentive segmentation module are depicted in Fig. \ref{fig:segmentation-module}.
%The segmentation network internally defines how the TD activities modulates the BU activities and the output is fused into the main processing pipeline of the segmentation network. 
% somewhere say this ...
Part (e) of Fig. \ref{fig:ssn-overall-modular} schematically illustrates the information flow from the BU network for feature encoding to the LSD for multi-instance and multi-scale label predictions. Attention initialization and TD network produce the selection patterns at multiple levels of the hierarchy. Lastly, the segmentation network produces the segmentation output maps through a number of parametric layers. This is the last stage of the inference phase and the segmentation output map is used to predict the category label of each pixel of the input image.

Fig. \ref{fig:ssn-overall-architecture} shows that the segmentation network has a number of processing layers. Each layer receives two inputs one from the corresponding BU layer and one from the corresponding TD layer. The inputs are transformed, fused, and up-sampled using a number of parametric building blocks at each layer. All of these block are necessary to form a representation given the inputs for the segmentation prediction.

%\textbf{Formulation:}
Once the two BU and TD passes are completed, Information is passed to
the segmentation network

\begin{equation}
	o=m(h_{i},g_{i};W_{seg}),
\end{equation}
in which $m=\{(b_{i},r_{i},p_{i},q_{i})\}_{i=0}^{M}$ consists of
$M$ segmentation layers, $W_{seg}$ is the set of segmentation weight
parameters, $h_{i}$ and $g_{i}$ are the hidden and gating activities at layer $i$ respectively, and $o$ is the segmentation output map which will be
used in the segmentation loss function in Sec. \ref{sec:seg:model:Multi-loss-Training}. 

The BU hidden activities $h_{i}$ and the TD gating activities $g_{i}$
have two different data distributions, one is densely active and the
other sparsely active due to the selective nature of the TD mechanisms. Therefore, the parametric
layers $o_{i}^{b}=b_{i}(b_{i}^{BU}(h_{i},W_{seg}^{BU}),b_{i}^{TD}(g_{i},W_{seg}^{TD}))$
are used to learn an appropriate transformation of the two inputs before the
TD modulation of BU activities. Three types of modulation are defined
for the fusion of the TD and BU activities: $b(u,v)=u\diamond v|\diamond\in\{\oplus,\odot,\ominus\}$,
$\oplus$ tensor summation, $\odot$ tensor multiplication, $\ominus$
tensor concatenation. 

After the modulation unit $b_i$, there is a parametric
layer $o_{i}^{r}=r_{i}(o_{i}^{b};W_{seg}^{r})$, a concatenation layer $o_{i}^{p}=p_{i}(o_{i}^{r},o_{i-1})\,|\,p(u,v)=u\ominus v$ to fuse the incoming information at layer $i$ into the information at layer $i+1$ in the segmentation pipeline,
and lastly another parametric layer $o_{i}=q_{i}(o_{i}^{p};W_{seg}^{q})$
followed by a spatial up-sampling layer. The modulation type $\diamond$,
the number of segmentation levels $M$, and other hyper-parameters
are cross-validated in the experimental evaluation in Sec. \ref{sec:obj_seg:exp}. All these building blocks at a segmentation layer are illustrated in Fig. \ref{fig:segmentation-module} with details such as the information flow from one computational unit to another one, the name and the computation type of each unit.
%demonstrates all the computational units of the segmentation network at each level.

\begin{table}
	\resizebox{0.60\columnwidth}{!}{
		
		\begin{tabular}{l|c|ccc}
			\hline 
			\multicolumn{2}{c|}{Network} & level 1 & level 2 & level 3\tabularnewline
			\hline 
			\multirow{5}{*}{AlexNet} & name & conv3\_3 & pool2 & pool1\tabularnewline
			\cline{2-5} 
			& input & 256 & 192 & 64\tabularnewline
			& b & 128 & 96 & 48\tabularnewline
			& r & 128 & 96 & 48\tabularnewline
			& q & 96 & 48 & 32\tabularnewline
			\hline 
			\multirow{5}{*}{VGG} & name & conv5\_3 & pool4 & pool3\tabularnewline
			\cline{2-5} 
			& input & 512 & 512 & 256\tabularnewline
			& b & 384 & 256 & 128\tabularnewline
			& r & 384 & 256 & 128\tabularnewline
			& q & 256 & 128 & 64\tabularnewline
			\hline 
		\end{tabular}
		
	}
	\centering{}\caption{The output channel size of computational units in the segmentation layers is given for AlexNet and VGG at three different levels. b, r, q are the units defined in \ref{sec:seg:model:Segmentation-Prediction}.\label{tab-sem-seg-channel}}
\end{table}

\subsection{LSD Pre-training}
\label{sec:seg:model:LSD-Pre-training}

%taken from the LSD math section
The learned feature representation of the pre-trained \convnet loaded on the BU network needs to be adapted to the new data domain and visual task of semantic segmentation.
The BU network parameters are initialized by the parameters of a convolutional neural network pre-trained on the Imagenet \cite{Russakovsky2015} dataset for the task of object classification.
The input images in Imagenet contain one single instance of an object category. 
The label prediction output of the classification network has the size of
%The loss is also measured using the network label prediction output maps of
$K\times1\times1$ since the input image size is smaller than the overall network receptive field. As a result, the loss function is measured only on one set of label predictions for all categories.
SSN, on the other hand, deals with input images that may contain multiple instances of the pre-defined labeled categories. Additionally, due to the larger size of the input images, the network returns output maps with spatial sizes greater than 1.
This shift of domain and task requires a preliminary stage of fine-tuning of the BU network using the loss function defined on the LSD module. 

The BU network needs to learn to accommodate for the new task and domain requirements in this first stage of the learning phase as is depicted in part (f) of Fig. \ref{fig:ssn-overall-modular}. We define the objective function $\mathcal{L_{LSD}}(\hat{p},t)=\frac{1}{N_{D}}\sum^{A}_{i}\mathcal{L}_{D}(\hat{p}_{i},t_{i})$ on top of the LSD module to minimize the loss of the prediction distribution $\hat{p}$ for the target label $t$. $\hat{p}$ is computed using the Softmax transfer function from the output score maps $s$. The SGD optimization algorithm updates the BU and LSD weight parameters using the gradient signals computed by the backpropagation algorithm.

We will fine-tune the pre-trained BU network following a class-specific
approach inspired from the SSD object detection model \cite{liu2016ssd}.
For an input image $x_i$, there is a set of bounding box annotations $y^B_i$ in the dataset $D$ which needs to be utilized to generate appropriate target labels for the training of the LSD module. 

Unlike the dominant object detection models, SSN only needs to loosely know if a LSD output unit is required for activating a TD selection mechanism or not. Therefore, it is not needed in SSN to have multiple scale, aspect ratio, and offset value predictions at each output unit. The LSD module in SSN only requires to output category label predictions which are harnessed later on by the attention initialization unit for the activation of a number of TD selection mechanisms. Part (g) of Fig. \ref{fig:ssn-overall-modular} demonstrates the role of the anchor processing unit for the computation of the LSD loss function.

%We speculate equipping the BU network to loosely learn how to predict category labels at different spatial locations improves the accuracy of the TD selection and ultimately the final segmentation predictions.
%Additionally, $s$ is the source of the TD selection initialization.
%Consequently, LSD needs to be robust enough to visual inter and intra-category variations. 

%\textbf{Formulation:}
For the training of LSD, we generate target labels $t=\{t_{i}\}_{i=0}^{A}$
to fine-tune the weight parameters of the BU network and the LSD module. We follow an anchor generation approach commonly practiced for object detection in \cite{liu2016ssd,ren2015faster}.
We define an anchor $a_{ij}$ for each LSD output unit
$j$ at the group layer $i$, and calculate its box coordinates from
the total receptive field size at that level. We then propose to set the target labels according to the following policy:

\[
t_{ij}=\begin{cases}
k & IoU(a_{ij},y^B_{k})>\theta^{pos}\\
0 & IoU(a_{ij},y^B_{k})<\theta^{neg}
\end{cases},
\]
where $k\in\{1,\dots K\}$.
the target $t_{ij}$ is assigned to a category label for which the Intersection-over-Union (IoU) metric of the corresponding anchor box $a_{ij}$ with a ground truth box $g_{k}$ for the category $k$ is over the positive threshold value $\theta^{[pos]}$. It is zero if the IoU metric is below the the negative threshold value $\theta^{[neg]}$. We always ensure that there is at least one anchor set for a ground truth bounding box.

\subsection{Multi-loss Training}
\label{sec:seg:model:Multi-loss-Training}

%\textbf{Definition:}
SSN training using the multi-loss function is the last stage of the learning phase as illustrated in part (h) of Fig. \ref{fig:ssn-overall-modular}. The BU network and LSD module parameters are loaded with the converged set of parameters in the LSD pre-training stage. The optimization algorithm considers two loss functions at the opposite ends of the visual hierarchy. The SSN parameters of the converged model is used in the inference phase for the segmentation prediction of unknown test images.

%\textbf{Formulation:}
SSN has output layers at the two ends of the visual hierarchy. The
first receives the LSD score maps $s$ as inputs at the
top of the visual hierarchy and outputs a discrete probability distribution
$\hat{p}(s)=\{\text{Softmax}(s_{i0},\dots,s_{iK-1})\}_{i=0}^{A}$
over $K$ categories including the background. On the other side at
the bottom of the hierarchy, segmentation output map $o$ is fed into
another output layer and returns similarly a discrete probability
distribution $\tilde{p}(d)=\{\text{Softmax}(d_{i0},\dots,d_{iK-1})\}_{i=0}^{H\times W}$,
where $H$ and $W$ are the height and width of the input image.
The overall multi-loss objective function is a weighted sum of the
LSD loss $\mathcal{L}_{D}$ and the segmentation
loss $\mathcal{L}_{S}$ terms

\begin{equation}
	\mathcal{L}_{Total}(\hat{p},t,\tilde{p},y)=\frac{1}{N_{D}}\sum_{i}\mathcal{L}_{D}(\hat{p}_{i},t_{i})+\alpha\frac{1}{N_{S}}\sum_{i}\mathcal{L}_{S}(\tilde{p_{i}},y_{i}),
\end{equation}
in which both of the loss functions $\mathcal{L}_{D}$ and $\mathcal{L}_{S}$
are defined using the element-wise negative log likelihood (NLL) function
for the true target labels $t_{i}$ and the segmentation mask $y^S_{i}$ respectively, and
$N_{D}=A$ and $N_{S}=H\times W$.

\section{Experimental Results}
\label{sec:exp}
We evaluate the performance of SSN on object segmentation to support
the role of a top-down selection mechanism in neural network
approaches. Semantic segmentation is the task that is defined to predict
segmentation masks of a pre-defined number of semantic categories
\cite{chen2018deeplab,everingham2015pascal,LinMaireBelongieEtAl2014,cordts2016cityscapes}. Challenging benchmark datasets such as PASCAL VOC \cite{Everingham2014},
the Cambridge-driving Labeled Video (CamVid) \cite{brostow2009semantic},
and Horse-Cow Parsing \cite{wang2015semantic} datasets are used for
experimental evaluation of SSN. 

SSN is implemented using PyTorch\footnote{\href{https://pytorch.org/}{https://pytorch.org/}}
\cite{paszke2017automatic}, an open source deep learning platform
which is well-known for its automatic differentiation engine. The TD pass
is integrated into the main implementation using the open-source CUDA
library provided by\footnote{\href{https://github.com/mbiparva/stnet-object-localization}{https://github.com/mbiparva/stnet-object-localization}}
\stn \cite{biparva2017stnet}. LSD pre-training begins with the Imagenet
pre-trained models provided by the PyTorch Model Zoo repository. The core architecture of the BU network in all of our experiments are defined based on either AlexNet \cite{krizhevsky2012imagenet} or VggNet \cite{simonyan2013deep} networks.

\begin{table}
	\resizebox{0.8\columnwidth}{!}{
		
		\begin{tabular}{l|cc|cc}
			\hline 
			\multirow{2}{*}{Model} & \multicolumn{2}{c|}{Parallel} & \multicolumn{2}{c}{Sequential}\tabularnewline
			\cline{2-5} 
			& m Accuracy & m IoU & m Accuracy & m IoU\tabularnewline
			\hline 
			AlexNet & 54.3 & 39.8 & 52.6 & 39.1\tabularnewline
			VGGNet & 66.7 & 55.6 & 67.1 & 53.6\tabularnewline
			\hline 
		\end{tabular}
		
	}
	\centering{}\caption{Parallel and Sequential LSD performance results on the Pascal VOC
		2012 validation set once the BU network is fine-tuned on the extended
		Pascal dataset.\label{tab-sem-seg-lsd}}
\end{table}

\subsection{Implementation Details\label{subsec:id}}
We provide details on the three processing modules of SSN for the
experimental evaluation on semantic segmentation benchmark dataset:
1) the BU representation is the core visual hierarchy that consists
of multiple parametric layers, 2) TD selection generates attentional
traces throughout the visual hierarchy with the most effective feature
modulation capability, and 3) attentive segmentation modulates hidden
activities with gating activities at different levels and merges them
into a unified representation with gradual up-sampling for the final
segmentation prediction. 

\textbf{Bottom-Up Feature Encoding:}
In this work, we use AlexNet \cite{krizhevsky2012imagenet} and VGG-16
\cite{simonyan2013deep} network architectures to define the ordering, connectivity and parametrization of the BU layers. The former has a smaller number of layers while the latter has more layers with parameters. In this work, we use AlexNet as a proof-of-concept network due to the simplicity of the architecture and fewer number of parameters. We first begin experimenting with SSN using AlexNet and then later extent to VGG to validate the experimental evaluation results and demonstrate the generalization to a larger network.
As illustrated in Fig. \ref{fig:ssn-bu-archs}, the BU network based on the two architectures has the following set of layers respectively: \{conv1, pool1, conv2, pool2, conv3\_1, conv3\_2, conv3\_3\} and \{conv1\_1, conv1\_2, pool1, conv2\_1, conv2\_2, pool2, conv3\_1, conv3\_2, conv3\_3, pool3, conv4\_1, conv4\_2, conv4\_3, pool4, conv5\_1, conv5\_2, conv5\_3\}. The feature channel of the hidden maps at each layer are given in Fig. \ref{fig:ssn-bu-archs}.
The last layer of the BU network is Conv3\_3 and Conv5\_3 in the two architectures respectively. For the input image size $320\times320$, the hidden activity output $h$ has the spatial size $20\times20$ in the both architectures. 

\textbf{Loose Spatial Detection:}
LSD has three groups of layers each of which has a set of intermediate layers followed by a final prediction layer. We define the number of groups $C$ to be three as it is sufficient to fully cover the small, medium, and large category objects.
The three sets of intermediate layers respectively consist of \{c1x1, c1x1\}, \{c3x3-p2-d2, c1x1, c3x3-p1\}, and \{m3x3-s2,
c3x3-p2-d2, c1x1, c3x3-p1\}. c, s, p, d, m stands for a convolutional layer,
stride, padding, dilation, and max pooling values respectively. c3x3-s2-p2-d2
defines a convolutional layer with the kernel size 3x3, stride 2,
marginal padding 2, and dilation 2. For the sake of brevity, the default values of s1, p0, and d1, are ignored.
Both of the sequential and parallel LSD predictors have layers with the same set of settings and hyperparameters. They only differ in terms of the ordering and connectivity of the groups with respect to each other. The input to LSD is taken from the intermediate layer conv5 and conv5\_3 in AlexNet and VGG respectively. The output score map $s$ is used by the attention initialization unit in the inference phase and the LSD loss function in the learning phase.

\textbf{Top-Down Selection:}
The TD selection is computed at each layer of the visual hierarchy,
gating activities are determined and the selection is passed to the
next layer, which is below the current one. Layer by layer selection
is executed until a particular stopping layer is met. The \textit{pool1} and \textit{pool3} layers are the stopping layer $V$ in AlexNet- and VGG-based BU networks respectively.

Unlike STNet model, SSN does not have any fully-connected layers. All the fully-connected layers are replaced with the convolutional layers that have kernel size $1\times1$. We refer to $1\times1$ convolutional layers as collapsed convolutional layers hereafter. Collapsed convolutional layers are technically fully-connected layers that are applied over two-dimensional feature maps rather than a one-dimensional feature vector. To address this requirement, we implemented the TD selection stages for collapsed convolutional layers using the stages for fully-connected layers. 

The TD pass has one hyperparameter at each layer in the second selection stage while the first and the last stages do not have any hyperparameter.
The second stage of the TD pass in STNet for the collapsed convolutional layers has a statistically-motivated thresholding value that determines how tight or loose the selection is. We replace it with the Winner-Take-All (WTA) mechanism since the nature of the object segmentation in this work is different from the object localization STNet was developed for. For the typical convolutional layers, there is a fusion factor that determines how much emphasis should be given to the spatial contiguity or the total activity strength. In STNet, it is called $\alpha$. We experimentally choose to set $\alpha=0.2$.

\textbf{Segmentation Prediction:}
The output channel size of each computation block at a segmentation
layer is given in Table \ref{tab-sem-seg-channel}. All of the convolutional
layers in the segmentation network have the kernel size of 3x3 with stride 1, padding 1, and dilation 0 unless otherwise mentioned.
%\textbf{Linking Module:}
As illustrated in Fig. \ref{fig:segmentation-module}, $h_{i}$ and $g_{i}$ are the two inputs to the segmentation layer $i$. They are first fed into $b^{BU}$ or $b^{TD}$, which are $3\times3$ convolutional layers, to reduce the feature channel size for a compact feature representation.
Next, the outputs are fused into one feature tensor by the modulation
operation $b_i$ which can be tensor concatenation, addition, or multiplication. The output of the modulation unit is sent into the convolutional layer $r_{i}$
to further reduce the feature redundancy before getting fused into the main segmentation pipeline.
%\textbf{Up-Sampling Module:} 
The feature tensor $o^r_i$ at this point is merged into the main segmentation pipeline using the feature concatenation unit $p_{i}$ along the feature channel dimension. 
The concatenation of $o^r_i$ is with the segmentation activities $o_{i+1}$ passed from the segmentation layer $i+1$. 
At the first level, the LSD label predictions are
used for the concatenation. This helps the error gradient signals computed using the segmentation loss function to reach to the LSD module and consequently flow downward through the BU visual hierarchy.
This brings faster optimization convergence during the training phase.
Next, the convolutional layer $q_{l}$ reduce the feature channel size further while the output spatial size is increased using a bilinear up-sampling layer by the factor of $2$. $o_i$ is the output of the segmentation layer $i$ and the input tensor for the next segmentation layer $i-1$.
%and is sent to the next level in the attentive segmentation pipeline. 
The details on the number of segmentation levels and the output channel size of each computational block is given in the Table \ref{tab-sem-seg-channel}. 
%We follow this layer definitions in the experimental evaluation in Sec. \ref{sec:obj_seg:exp}.

\textbf{LSD Pre-training:}
Following the experimental setting in \cite{ren2015faster} and preliminary experimental results, we choose to set $\theta_{pos}=0.5$ and $\theta_{neg}=0.3$ in the experimental evaluation phase. 
A target label has the category label of the box overlapping with the corresponding anchor if the IoU value of the box with the anchor is above $\theta_{pos}$. It has the background label zero if the IoU of the two is below $\theta_{neg}$. 
We set the otherwise to the don't-care label value $255$. Once all of the target
labels are set for the bounding boxes annotations of the mini-batch samples, we randomly keep a maximum number of 128 target labels per mini-batch samples such that the ratio between the negatives and positives is at most 1:3 while
the rest are set to $255$. 
We set the element-wise cross-entropy loss function $\mathcal{L_{LSD}}$ to exclude the loss terms of the output units for which the target labels are set to $255$.
%The optimization algorithm ignores including the loss values of the units with the target label $255$.

%\textbf{LSD Training Procedure:} 
Given an input image, the LSD output units are computed using a feedforward pass and the target labels are determined using the ground truth bounding boxes. We fine-tune the parameters of the BU network and the LSD module using the SGD optimizer with the initial learning rate $10^{-3}$, momentum $0.9$, weight decay $0.0005$, and batch size $4$ for $15$ epochs. We evaluate the performance of LSD using similar segmentation metrics namely the mean pixel and the mean IoU performance metrics. Once the LSD pre-training is converged, we train the SSN model using the multi-loss function described in Sec. \ref{sec:seg:model:Multi-loss-Training}.

\textbf{Multi-loss Training:}
%\textbf{Segmentaiton Prediction Module:} 
The output of the last up-sampling layer is the input of the final segmentation prediction module.
This module simply consists of one 3x3 and one 1x1 convolutional layers:
the former keeps the feature channel size intact, and the other reduces
it to the number of category labels $K$. The output of this module
is the confidence scores used by a Softmax layer to produce multinomial
probability values. Similar to LSD pre-training procedure, we use an element-wise
cross-entropy loss function to optimize the set of all of the weight
parameters of SSN $W_{SSN}=\{W_{seg},W_{LSD},W_{BU}\}$.
%multinomial logistic regression

%\textbf{Error Gradient Propagation:} 
Since we have a multi-loss objective function, the error gradient signals are propagated from the two ends of SSN: the first loss term propagates error signals from the top of the visual hierarchy all the way to the input layer, while the second loss propagates the error signals from the bottom of the visual hierarchy. The error signals measured from the segmentation loss propagate into the BU network according to the modulatory patterns generated by the TD gating activities. This has an important impact on the underlying representation of the BU network. While the LSD loss keeps the representation fidelity of the BU network, the second loss term updates the parameters of the hierarchical transformation for a more robust and adapted segmentation prediction.

\begin{table}
	\resizebox{0.7\columnwidth}{!}{
		
		\begin{tabular}{l|cc}
			\hline 
			\multirow{1}{*}{SSN Variant} & mean Accuracy & mean IoU\tabularnewline
			\hline 
			SSN-GT & 53.7 & 41.9\tabularnewline
			SSN-MAX & 51.9 & 40.4\tabularnewline
			SSN-THD & 53.5 & 41.4\tabularnewline
			\hline 
			SSN-THD-CAT & 50.6 & 40.2\tabularnewline
			SSN-THD-ADD & 53.5 & 41.4\tabularnewline
			SSN-THD-MUL & 54.1 & 42.1\tabularnewline
			\hline 
		\end{tabular}
		
	}
	\centering{}\caption{Comparison of different variants of SSN using AlexNet on PASCAL VOC
		valid 2012. We use mean pixel accuracy and mean IoU metrics to report
		the performance. CAT, MAX, THD, ADD, and MUL stands for concatenation,
		top-1, and thresholding, additive, and multiplicative.\label{tab-sem-seg-variants}}
\end{table}

\begin{table}
	\resizebox{0.80\columnwidth}{!}{
		
		\begin{tabular}{l|l|cc}
			\hline 
			Network & \multirow{1}{*}{Model} & Mean Accuracy & Mean IoU\tabularnewline
			\hline 
			\multirow{4}{*}{AlexNet} & FCN & - & 39.8\tabularnewline
			& SSN & 54.1 & 42.1\tabularnewline
			\cline{2-4} 
			& FCN++ & - & 48.0\tabularnewline
			& SSN++ & 55.8 & 49.4\tabularnewline
			\hline 
			\multirow{4}{*}{VGGNet} & FCN & - & 56.0\tabularnewline
			& SSN & 64.6 & 58.7\tabularnewline
			\cline{2-4} 
			& FCN++ & 75.9 & 62.7\tabularnewline
			& SSN++ & 76.8 & 64.3\tabularnewline
			\hline 
		\end{tabular}
		
	}
	\centering{}\caption{Comparison of SSN with the baseline model on PASCAL VOC validation
		set using mean IoU metric. SSN++ is trained on the extended training
		set.\label{tab-sem-seg-valid}}
\end{table}

\begin{table}
	\resizebox{0.60\columnwidth}{!}{
		
		\begin{tabular}{l|c}
			\hline 
			Method & mean IoU (\%)\tabularnewline
			\hline 
			FCN \cite{long2015fully} & 62.7\tabularnewline
			DeepLab \cite{chen2016semantic} & 64.2\tabularnewline
			G-FRNet \cite{islam2017gated} & 68.7\tabularnewline			
			DeepLab-ASPP \cite{chen2018deeplab} & 68.9\tabularnewline
			\hline 
			SSN(ours) & 64.3\tabularnewline
			\hline 
		\end{tabular}
		
	}
	\centering{}\caption{Comparison of SSN with the state-of-the-art on PASCAL VOC 2012 valid set. All methods use VGGNet as the backbone network. \label{tab-sem-seg-pascal-sota}}
\end{table}

\subsection{Semantic Segmentation\label{subsec:ss}}

Dense image labeling such as semantic segmentation requires
pixel-level predictions. In this section, we first measure the performance
of a variety of SSN configurations on predicting accurately object
segmentation of various categories of the input images. Later, we
provide the comparison with the baseline model and discuss the aspects
the superiority originates from. 

\textbf{Dataset and Evaluation:} A popular and challenging benchmark
dataset for semantic segmentation is PASCAL VOC 2012 \cite{Everingham2014,everingham2010pascal}.
The dataset contains 1464 training and 1449 validation sample images.
Each image may have a number of instances of 21 object categories
(including the background category). To enable comparisons with previous
works, the training set is expanded with extra labeled data \cite{hariharan2011SBD}.
We measure the segmentation performance using the mean pixel accuracy
and mean IoU metrics \cite{long2015fully}. In the training phase,
we resize sample images to have the smallest side of 320 and then take a random
crop of size 320x320. In the evaluation phase, we resize images to have
the largest side of 320 and pad the smallest side of the RGB image
the mean pixel values and the segmentation mask with don't care pixel
values. We follow this strategy for all the experiments in this work.

\textbf{Quantitative Results:} We first fine-tune the Imagenet pre-trained
AlexNet and VggNet networks using the LSD training protocol on the
extended Pascal VOC 2012 dataset. The performance results are given
in Table \ref{tab-sem-seg-lsd} using the two evaluation metrics.
The performance metric results using the parallel LSD outperforms
the sequential one for both of the \convnet model. Since the problem
in LSD formulation is merely a classification task, the parallel LSD
benefits from the separation and independence of label predictors
and consequently survives over-fitting to the training set. We stick
to the parallel LSD configuration for the rest of experimental evaluation.

We report the performance of different setups of SSN using the AlexNet
architecture in Table \ref{tab-sem-seg-variants}. First, we measure
the effect of the attention signal initialization strategy on the
SSN performance. Apparently, the GT initialization strategy is superior
over the other two since it minimizes redundancy and noise interference
imposed by TD selection from false alarm units. Mean IoU of 40.4 in
the max strategy increases to 41.4 in the thresholding strategy. This
indicates that the false alarm interference is reduced by thresholding
units with prediction confidence below the $\theta^{attention}$ of
0.90. 

Next, we investigate the effect of the modulation types of the segmentation
network. The selective nature of the TD gating activities is supported
since the multiplicative modulation outperforms the other two types.
This is reminiscent of the surround suppression phenomenon in human vision predicted in the Selective Tuning model of
visual attention \cite{tsotsos2011computational}. Using the multiplicative
modulation, BU hidden units that are not selected by the TD gating
units do not participate in the computation of the segmentation network
and therefore information flow is blocked or reduced in such units.
This underlines how a hierarchical selective mechanism can dynamically
suppress redundancy in the visual representation and lead to more
robust task predictions. We use the thresholding initialization strategy
and multiplicative modulation for the rest of experiments.

\textbf{Comparison with the baseline:} The best variant of SSN is
trained on PASCAL VOC 2012 training set and the extra data of \cite{hariharan2011SBD}
using both of the network architectures. The performance is reported
on the PASCAL VOC 2012 validation set in Table \ref{tab-sem-seg-valid}.
Samples in the validation set are excluded from the extended training
set. The SSN performance is compared with the baseline model of FCN \cite{long2015fully}.
The first goal is to introduce a well-established TD selection mechanism
and highlights the aspects that lead to improvements over the FCN
model. FCN introduced dense and parametric skip connections from early
layers for parametric up-sampling of label scores of a classifier for
segmentation. In all cases in Table \ref{tab-sem-seg-valid}, SSN
improves the mean IoU results over FCN for both of the architectures.

The quantitative results verify the modulatory role of the TD selection
mechanism in SSN. SSN benefits from an architecture that employs TD
selection to begin from high-level semantic layers and traverse to
intermediate-level feature representations. The TD traversal outputs
selection patterns at each layer that highlight important regions
and features along the visual hierarchy. The results emphasize that a systematic hierarchical gating of the information flow from the early feedforward layers
into the segmentation pipeline has positive impact on the evaluation
metrics. 

\textbf{Comparison with the state-of-the-art:} We compare the best segmentation performance of SSN on the PASCAL VOC 2012 validation set with the performance of the state-of-the-art methods in Table \ref{tab-sem-seg-pascal-sota}. SSN outperforms FCN by 1.6\% and is par with DeepLab which benefits from features such as multi-scale prediction method and Atrous (strided) convolutional layers with large field of views. The multi-scale method concatenate feature maps form the early layers with the network's last layer feature map. This helps the last feature maps to gain extra information to compromise for the lost of information imposed by the hierarchical feature encoding. SSN, on the other hand, does not benefit from multi-scale skip connections but rather relies on the TD selection mechanism to route through the network and highlights the important features for segmentation. G-FRN and DeepLab-ASPP are by approximately 4\% more accurate in predicting semantic segmentation in comparison with SSN. This is mainly due to the fact that SSN does not benefit from the features such as the stage-wise supervision in \cite{islam2017gated} and the Atrous Spatial Pyramid Pooling (ASPP) in \cite{chen2018deeplab}. The former provides strong supervision at multiple-levels of the feature hierarchy using different loss functions. This facilitates the error gradient propagation throughout the network hierarchy. ASPP employs parallel branches with different atrous rates at the top fully-connected layers to cover a wide range of filed of views. SSN uses none of these features and this explains the reason it falls behind these two methods.

\textbf{Additional Experimental Evaluation:} To further support experimentally
the role of the TD mechanism for the attentive segmentation framework
of SSN, we compare performance of SSN and FCN on two challenging datasets:
CamVid and Horse-Cow datasets. CamVid has 701 frames of urban driving
extracted from high resolution video recordings. Following \cite{kundu2016feature,badrinarayanan2017segnet,sturgess2009combining},
we consider 11 large semantic categories, down-sample images by a
factor of two (i.e. 480x360), and split them into the training (367),
validation (100), test sets (233). Horse-Cow part parsing dataset
contains semantic labeling of four body parts (head, leg, tail, body).
Following \cite{wang2015semantic}, we split the dataset into 294
training and 227 test images. We follow the resizing and padding protocol
introduced for PASCAL dataset.

The results in Table \ref{tab-sem-seg-additional} reveals the efficiency
of the TD selection to obtain segmentation robustness is consistent
across different datasets. SSN improves on the mean IoU metric values of FCN for
CamVid and Horse-Cow datasets and is on par on the mean accuracy metric values. 
%It is worth mentioning that the selective nature of SSN is better illustrated for datasets with higher number of semantic categories. It is indicated by the metric gap between the SSN and FCN for all three benchmark dataset. It is larger for PASCAL and gradually degrades for CamVid and Horse-Cow with smaller number of categories. 
We further compare the performance of SSN on these two benchmark datasets with DeepLab-LargeFOV \cite{chen2016semantic} and G-FRNet \cite{islam2017gated}. SSN outperforms DeepLab on the Horse-Cow and CamVid datasets. However, similar to the results of the PASCAL VOC dataset, SSN cannot compete with G-FRNet on these two datasets due to the extra design features G-FRNet benefits from. The evaluation results on these dataset are consistent with the previous experiments on the PASCAL VOC dataset. This reveals that the role of TD selection mechanism generalizes across different benchmark datasets for semantic segmentation. 

\begin{table}
	\resizebox{0.9\columnwidth}{!}{
		
		\begin{tabular}{l|l|cc}
			\hline 
			Network & \multirow{1}{*}{Model} & Mean Accuracy & Mean IoU\tabularnewline
			\hline 
			\multirow{2}{*}{CamVid} & FCN & 66.4 & 57.0\tabularnewline
			& DeepLab-LargeFOV* & - & 61.6\tabularnewline
			& G-FRNet\cite{islam2017gated} & - & 68.0\tabularnewline
%			& SSN & 68.9 & 60.7\tabularnewline
			& SSN & 73.9 & 64.7\tabularnewline
			\hline 
			\multirow{2}{*}{Horse-Cow} & FCN & 77.3 & 63.1\tabularnewline
			& DeepLab-LargeFOV* & - & 62.7\tabularnewline
			& G-FRNet\cite{islam2017gated} & - & 68.1\tabularnewline
			& SSN & 77.2 & 65.2\tabularnewline			
			\hline 
		\end{tabular}
		
	}
	\centering{}\caption{Comparison of SSN with the baseline and state-of-the-art on two additional segmentation benchmark datasets: CamVid and Horse-Cow. The results are reported on the test sets. Note that the DeepLab-LargeFOV* results are taken from\cite{islam2017gated}. 
		\label{tab-sem-seg-additional}}
\end{table}

\begin{figure}[]
	\centering
	\includegraphics[width=0.50\textwidth]{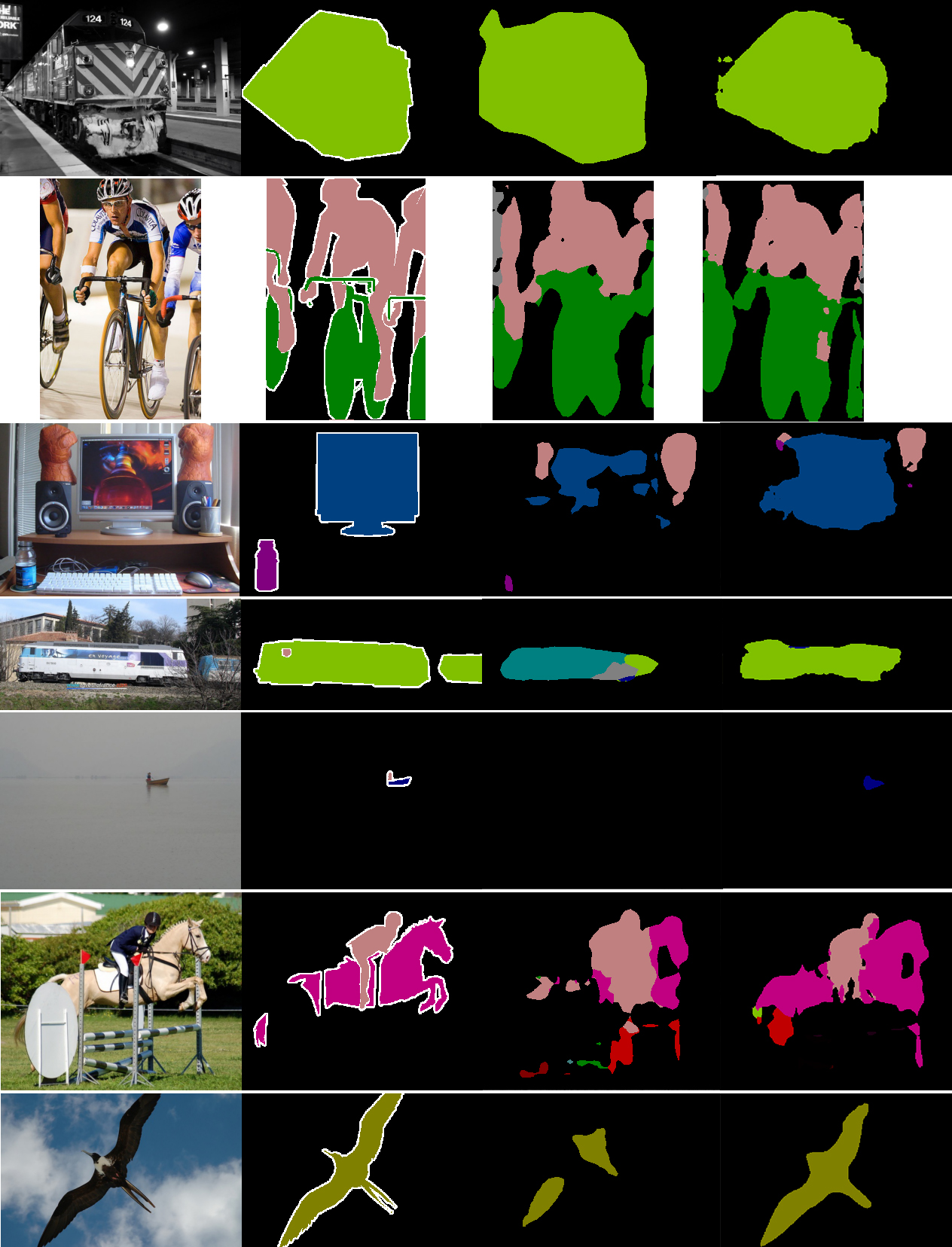}
	\caption{Comparison of the segmentation predictions of SSN with FCN on Pascal dataset. From left to right: RGB images, ground-truth, FCN predictions, SSN predictions.
		\label{fig:ssn-pascal-comp-qual}}
\end{figure}

\begin{figure}[]
	\centering
	\includegraphics[width=0.47\textwidth]{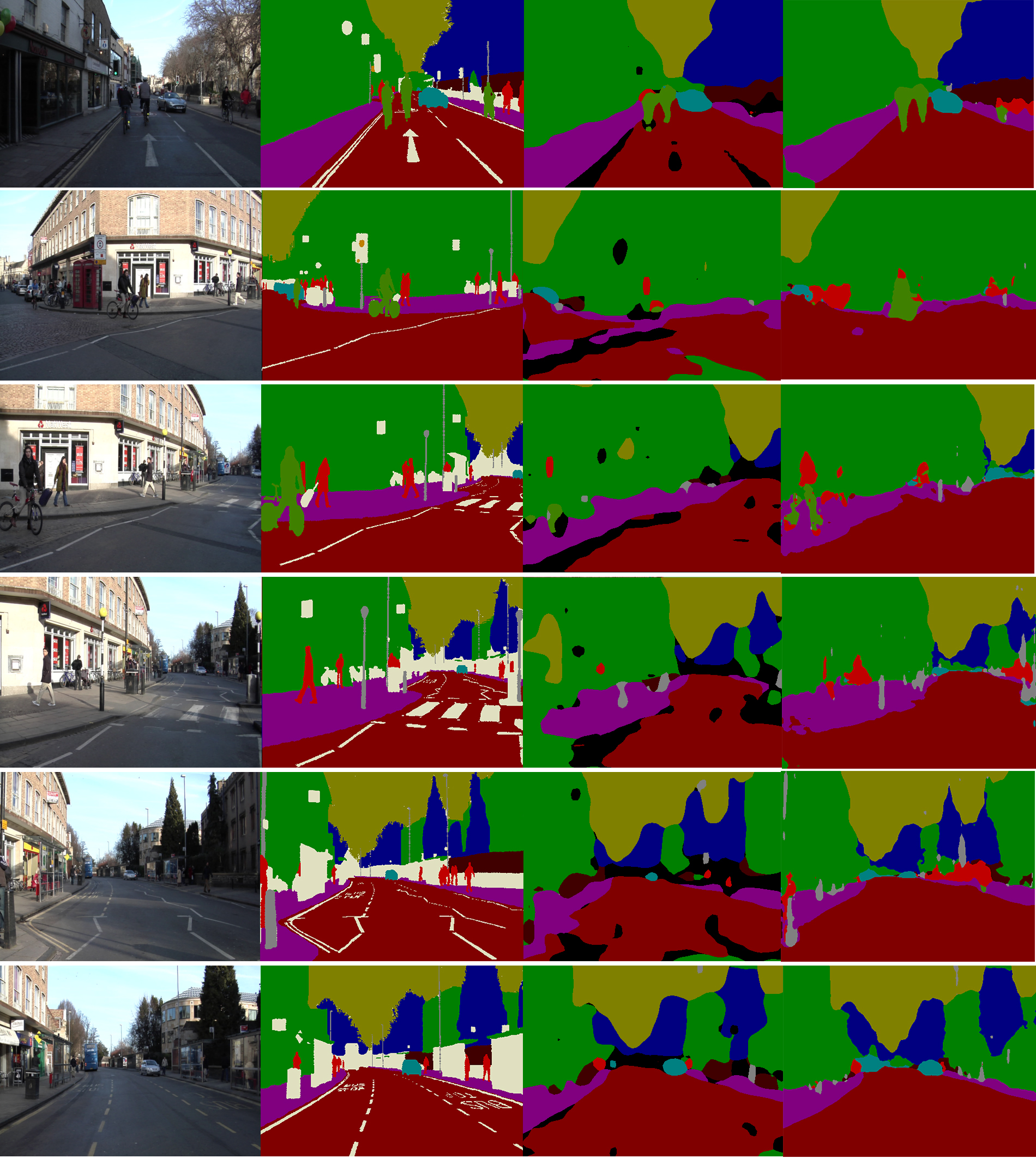}
	\caption{Comparison of the segmentation predictions of SSN with FCN on CamVid dataset. From left to right: RGB images, ground-truth, FCN predictions, SSN predictions.
		\label{fig:ssn-camvid-comp-qual}}
\end{figure}

\begin{figure}[]
	\centering
	\includegraphics[width=0.48\textwidth]{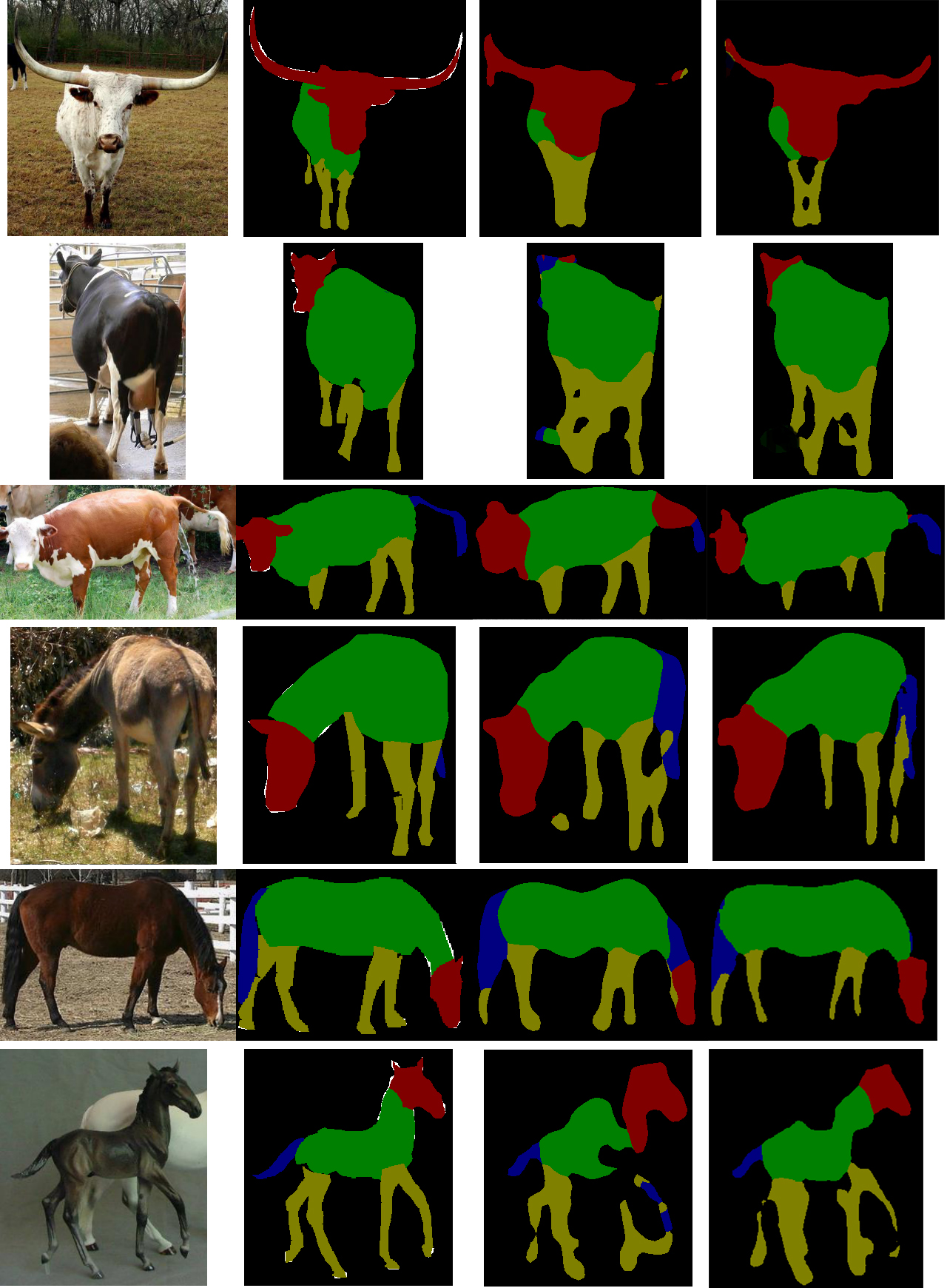}
	\caption{Comparison of the segmentation predictions of SSN with FCN on Horse-Cow part parsing dataset. From left to right: RGB images, ground-truth, FCN predictions, SSN predictions.
		\label{fig:ssn-horsecow-comp-qual}}
\end{figure}

\textbf{Qualitative Results:} We qualitatively compare SSN with the baseline FCN model on PASCAL, Cam-Vid, Horse-Cow part parsing datasets in figures \ref{fig:ssn-pascal-comp-qual}, \ref{fig:ssn-camvid-comp-qual}, \ref{fig:ssn-horsecow-comp-qual} respectively. The selectivity and modulatory role of the TD processing on the BU processing for the lateral connections is clearly depicted in cases the small object instances in the far distance are missed by FCN to be segmented successfully. Additionally, SSN is capable of predicting the shape of the segment masks and filling in the large regions in comparison to the FCN results. These aspects reveals the role of the TD selection in SSN to route relevant information from the BU pathway into the segmentation network.

\subsection{Ablation Studies}

We conduct further experiments to highlight aspects of SSN and emphasize
the critical role of the TD selection. First, we show the performance
deteriorates if either of the TD and BU inputs are blocked from feeding
into the segmentation network. In the upper part of Table \ref{tab-sem-seg-ablation},
performance drops from 42.1 for SSN with the segmentation network benefiting from the two inputs  (BU and TD activities) to 40.2 for only
BU hidden inputs and to 39.7 for only the TD gating inputs. Both inputs have complementary roles for the segmentation performance of SSN. The BU input benefits from the parametric distributed representation
while the TD input has a predictive selective characteristic. Once these two jointly are in place, the segmentation performance of SSN is superior to the either once individually used.

\begin{figure}[]
	\centering
	\includegraphics[width=0.48\textwidth]{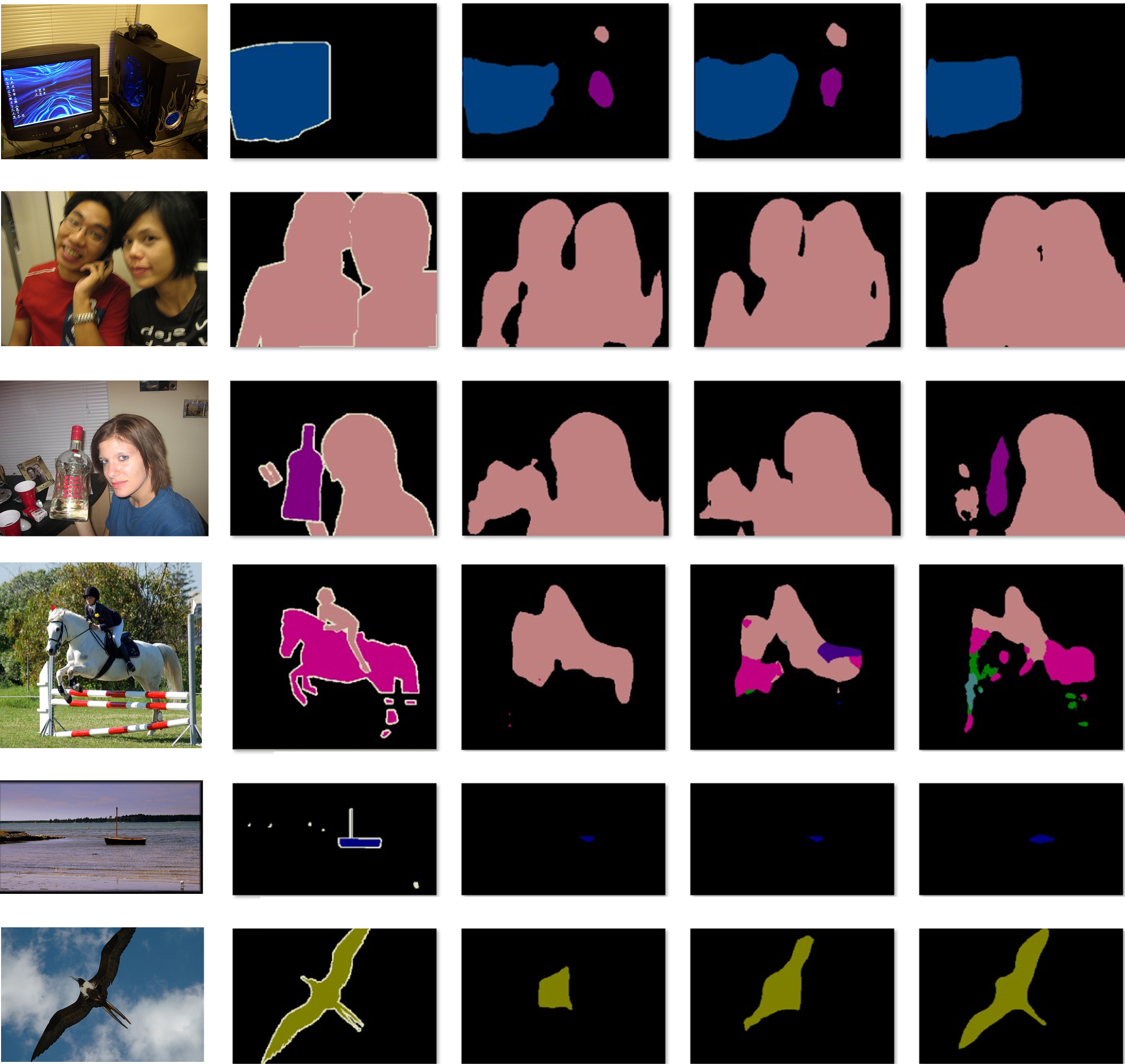}
	\caption{Demonstrating the role of the number of levels of TD and BU modulation on the segmentation prediction. From left to right: RGB images, ground-truth, SSN with 1 level of modulation, SSN with 2 levels of modulation, and SSN with 3 levels of modulation respectively.
		\label{fig:ssn-level-comp-qual}}
\end{figure}

Additionally, we emphasize the role of the number of levels of processing in the segmentation network on the SSN performance in the lower part of Table \ref{tab-sem-seg-ablation}. The segmentation pipeline with one level has
the least performance accuracy while as the number of levels increases,
the segmentation performance improves. This finding underlines that SSN benefits from the selectivity of the TD mechanisms on the high-level to the intermediate layer representations of the BU network for accurate object segmentation. The intermediate layers have fine details while the top layers have coarse structures. The relevance of the selectivity of the lower layer hidden activities imposed by the TD gating activities is supported by the results in this experiment. This is in line with the hypothesis that the TD selection process is capable of activating units in the spatial and channel dimensions for the modulation of the BU features for the dense pixel-level labeling task of semantic segmentation. In Fig. \ref{fig:ssn-level-comp-qual}, it is demonstrated that as the number of levels of modulation increases, the predictions become more accurate. False positive predictions are corrected and the shape of segmentation regions becomes more accurate. For instance, in the bird example, the shape of the segmentation for the bird becomes close the the ground truth once we have 3 levels of modulation in SSN. The qualitative results additionally highlight the modulatory role of the TD selection on the segmentation performance results. 
%The modulatory role of the TD mechanisms on the BU dense feature encoding is further verified in this round of experiments.

\begin{table}
	\resizebox{0.7\columnwidth}{!}{
	\begin{tabular}{l|cc}
		\hline 
		\multirow{1}{*}{Model} & mean Accuracy & mean IoU\tabularnewline
		\hline 
		SSN & 54.1 & 42.1\tabularnewline
		SSN-BU & 51.3 & 40.2\tabularnewline
		SSN-TD & 49.4 & 39.7\tabularnewline
		\hline 
		SSN-1 & 51.5 & 40.7\tabularnewline
		SSN-2 & 52.9 & 41.6\tabularnewline
		SSN-3 & 54.1 & 42.1\tabularnewline
	\end{tabular}
}
	\centering{}\caption{Ablation Studies on the TD modulatory role, the error signal propagation,
		number of gating layers into the segmentation pipeline using AlexNet
		on the Pascal VOC 2012 validation set.\label{tab-sem-seg-ablation}}
\end{table}

\begin{figure}[t]
	\includegraphics[width=0.9\linewidth]{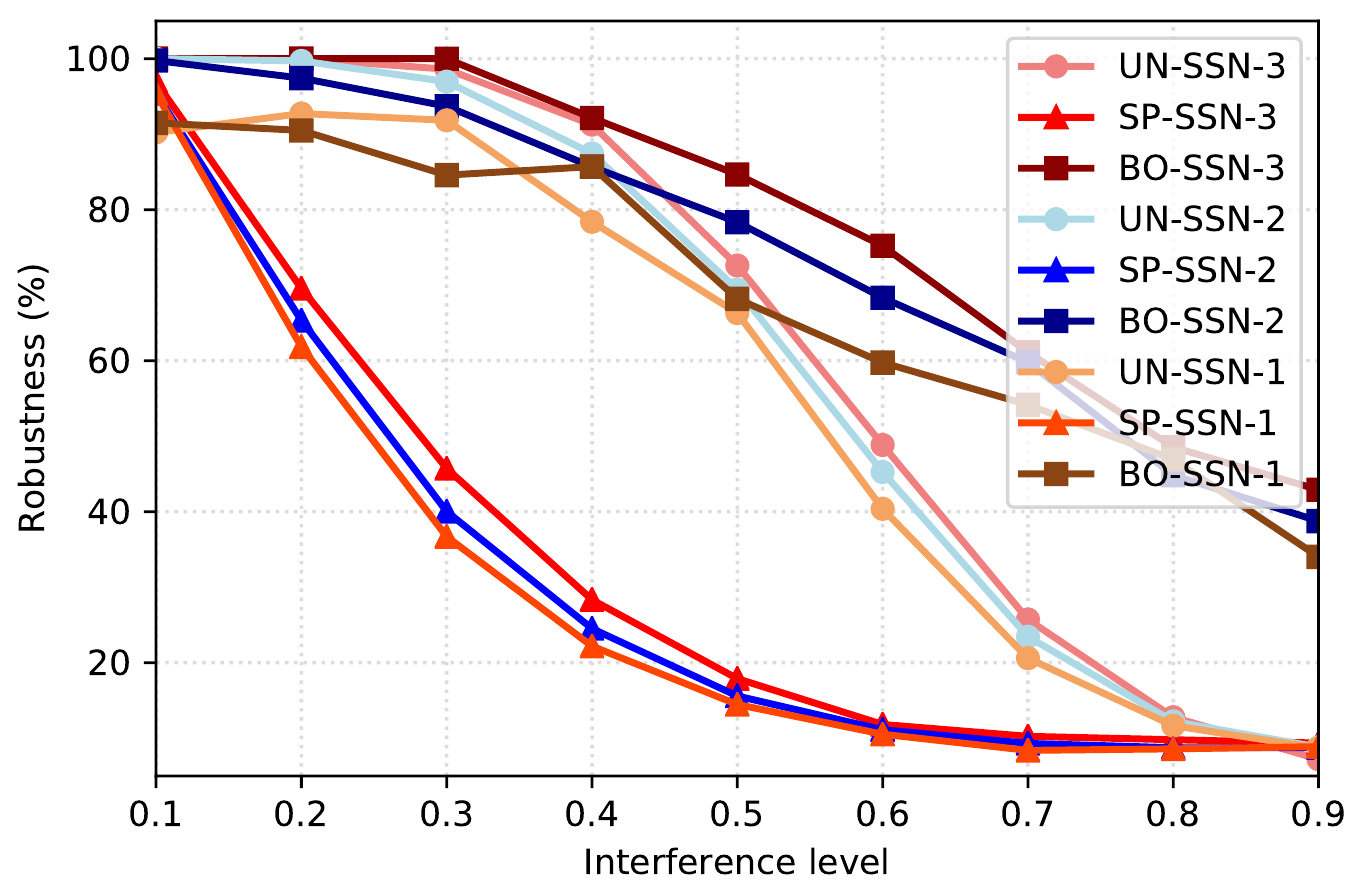}
	
	\caption{Robustness of SSN is measured at different interference levels ($\sigma$)
		for the uniform (UN), salt-pepper (SP), and box occlusion (BO) types.
		$\sigma$ determines the bandwidth of the uniform noise ($255\times\sigma$),
		the probability of having salt-pepper noise at a location, and the
		length of the occlusion box ($\sigma\times min(h_{image},w_{image})$)
		respectively. The modulatory role of the depth of the TD selection
		is demonstrated for SSN-k with inputs at k number of levels into the
		segmentation pipeline. \label{fig:robustness-noise-2}}
\end{figure}

\subsection{Noise Interference Robustness}

\begin{figure}[]
	\centering
	\includegraphics[width=0.48\textwidth]{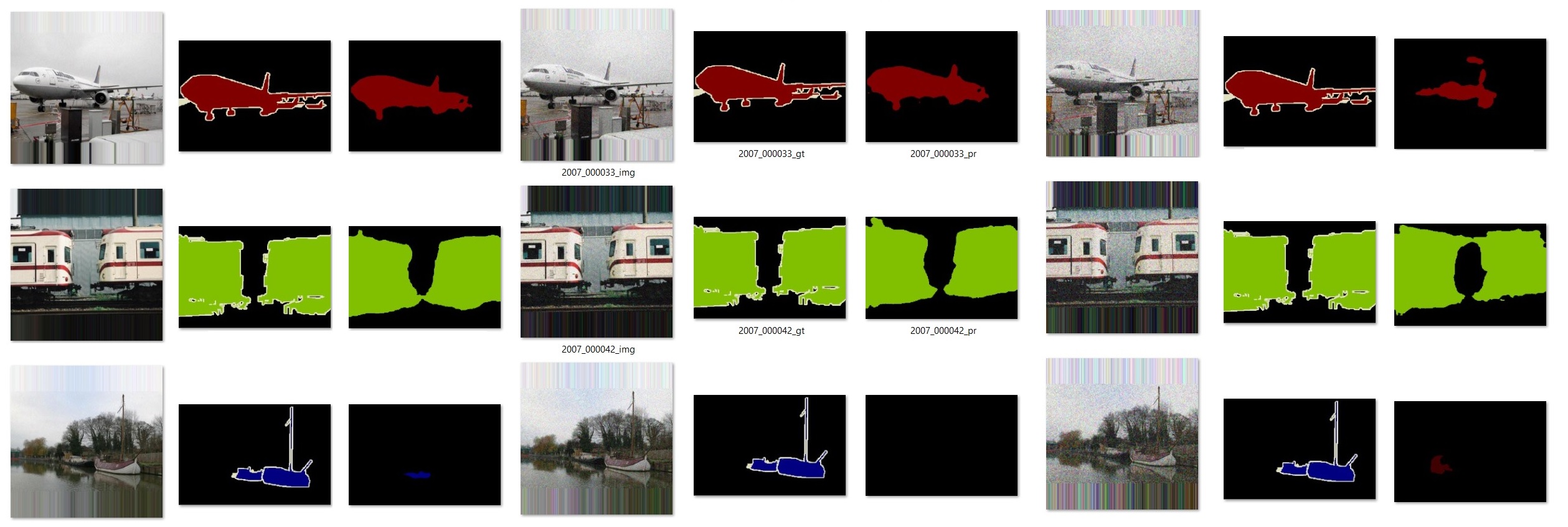}
	\caption{Three different levels of uniform noise is added to the RGB images. From left to right the noise level is 0.25, 0.45, 0.65 respectively.
		\label{fig:ssn-uni-comp-qual}}
\end{figure}

\begin{figure}[]
	\centering
	\includegraphics[width=0.48\textwidth]{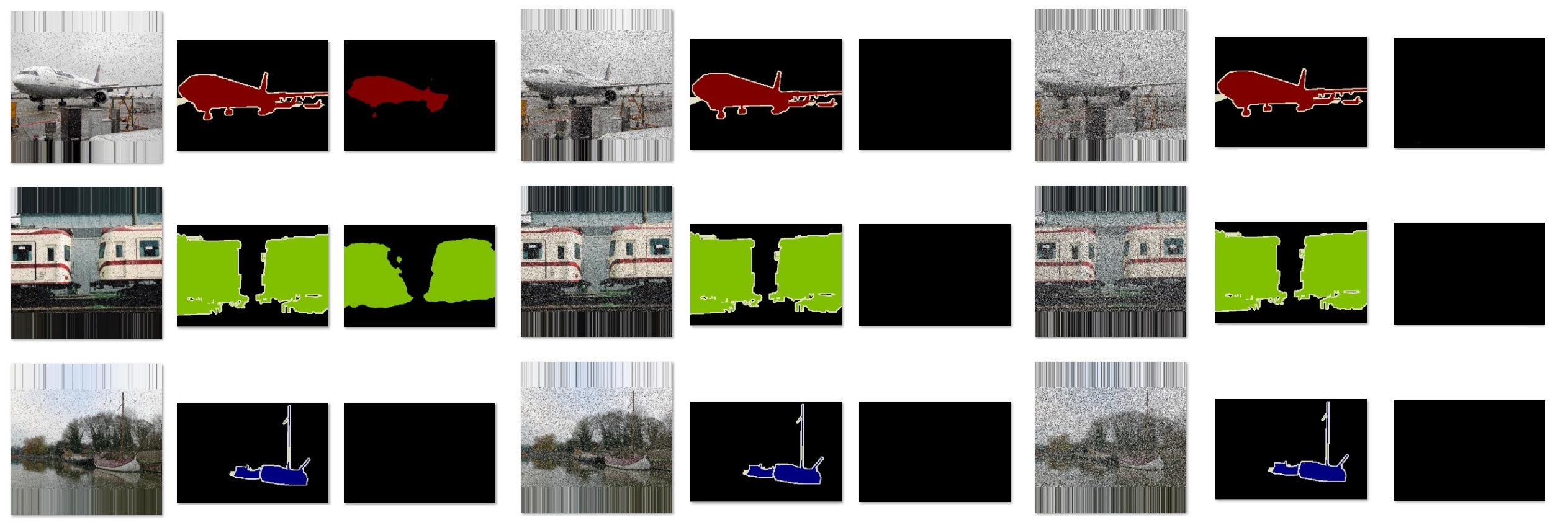}
	\caption{Three different levels of salt-pepper noise is added to the RGB images. From left to right the noise level is 0.25, 0.45, 0.65 respectively.
		\label{fig:ssn-sp-comp-qual}}
\end{figure}

\begin{figure}[]
	\centering
	\includegraphics[width=0.48\textwidth]{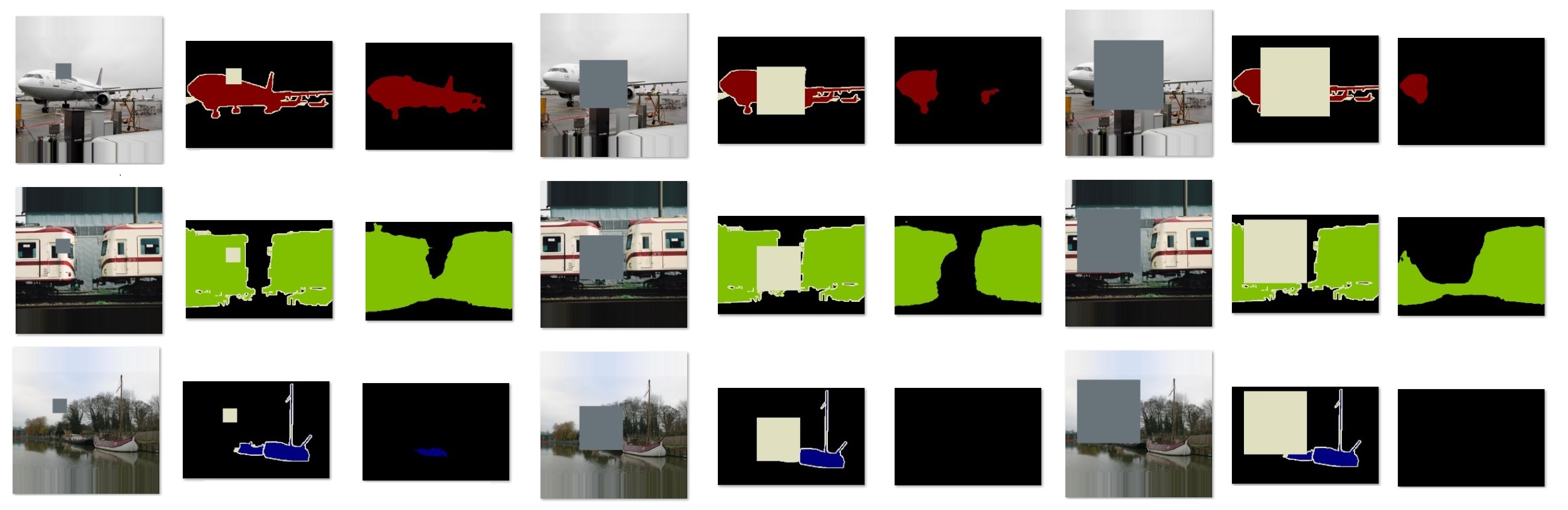}
	\caption{Three different levels of box-occlusion noise is added to the RGB images. From left to right the noise level is 0.25, 0.45, 0.65 respectively.
		\label{fig:ssn-box-comp-qual}}
\end{figure}

We test the robustness of SSN against two types of visual confusion:
noise interference and partial occlusion. We conduct experiments to
study the effects of the additive uniform noise, salt-and-pepper noise,
and box occlusion on SSN performance. Fig. \ref{fig:robustness-noise-2}
illustrates the role of the gating mechanisms derived by TD selection
at the early layers to gain increased level of robustness on perturbed
data samples. SSN with three levels of attentive segmentation obtains
the highest degree of robustness compared with smaller number of levels.
This is consistent across not only the two types of additive noise interference, but also partial box occlusion. Figures \ref{fig:ssn-uni-comp-qual}, \ref{fig:ssn-sp-comp-qual}, and \ref{fig:ssn-box-comp-qual} illustrates qualitatively how three levels of noise degrades the segmentation performance of SSN. The figures present the experimental setups that we used to study the noise interference robustness of SSN with 3 different modulation levels.

\section{Conclusion}
\label{sec:con}
The Top-Down selective attention is a well-known processing component of the human vision system. 
We introduce a unified Bottom-Up and Top-Down framework that
not only benefits from a feedforward representation but also a backward
selective modulation mechanism for the task of object segmentation in this work.
We define a parametric semantic controller to predict for the activation of TD
mechanisms at the top of the visual hierarchy. 
We demonstrate how the TD gating activities modulate the BU activities
for object segmentation through different stages of the information processing. The experimental evaluation results supports the role of the TD selection to improve the baseline performance results.

{\small
	\bibliographystyle{ieee}
	\bibliography{thesis_phd}
}

\end{document}